\documentclass[journal]{IEEEtran}
\usepackage{times}
\usepackage{latexsym}
\usepackage{tikz}
\usepackage{caption}

\usepackage[colorlinks=true, linkcolor=blue, citecolor=blue, urlcolor=black]{hyperref}

\usepackage[T1]{fontenc}
\usepackage[utf8]{inputenc}

\usepackage{microtype}

\usepackage{inconsolata}

\usepackage{graphicx}

\usepackage{makecell}
\usepackage{booktabs}
\usepackage{multirow}
\usepackage{algorithm}
\usepackage{algorithmic}
\usepackage{enumitem}
\usepackage{xcolor}
\usepackage{subfigure}
\usepackage{diagbox} 
\usepackage{amsmath}
\usepackage{amsfonts}
\usepackage{bbding}
\usepackage[numbers]{natbib}
\usepackage{pgfplots}
\pgfplotsset{compat=1.18}
\usepackage[edges]{forest}
\usepackage{tabularx}
\usepackage{makecell}

\usepackage[most,skins,theorems]{tcolorbox}
\tcbset{
  aibox/.style={
    width=\linewidth,
    top=8pt,
    bottom=4pt,
    colback=blue!6!white,
    colframe=black,
    colbacktitle=black,
    enhanced,
    center,
    attach boxed title to top left={yshift=-0.1in,xshift=0.15in},
    boxed title style={boxrule=0pt,colframe=white,},
  }
}
\newtcolorbox{AIbox}[2][]{aibox,title=#2,#1}

\ifCLASSINFOpdf
\else
\fi
\newcommand{\eg}{\emph{e.g.}}
\newcommand{\ie}{\emph{i.e.}}

\newif\ifshowcomments
\showcommentstrue   % 将该行注释掉来隐藏注释

\ifshowcomments
    \newcommand{\jc}[1]{{\color{purple} [jiachen: #1]}}
    
\else
    \newcommand{\jc}[1]{}
\fi

\ifshowcomments
    \newcommand{\mh}[1]{{\color{cyan} [zmh: #1]}}
    
\else
    \newcommand{\mh}[1]{}
\fi

\ifshowcomments
    \newcommand{\rrt}[1]{{\color{brown} [rrt: #1]}}
    
\else
    \newcommand{\rrt}[1]{}
\fi

\ifshowcomments
    \newcommand{\xx}[1]{{\color{orange} [xx: #1]}}
    
\else
    \newcommand{\xx}[1]{}
\fi

\ifshowcomments
    \newcommand{\ljh}[1]{{\color{red} [jh: #1]}}
    
\else
    \newcommand{\ljh}[1]{}
\fi

\begin{document}
%
% paper title
% Titles are generally capitalized except for words such as a, an, and, as,
% at, but, by, for, in, nor, of, on, or, the, to and up, which are usually
% not capitalized unless they are the first or last word of the title.
% Linebreaks \\ can be used within to get better formatting as desired.
% Do not put math or special symbols in the title.

\title{Evolutionary Perspectives on the Evaluation of LLM-Based AI Agents: A Comprehensive Survey}
%
%
% author names and IEEE memberships
% note positions of commas and nonbreaking spaces ( ~ ) LaTeX will not break
% a structure at a ~ so this keeps an author's name from being broken across
% two lines.
% use \thanks{} to gain access to the first footnote area
% a separate \thanks must be used for each paragraph as LaTeX2e's \thanks
% was not built to handle multiple paragraphs
%

\author{Jiachen~Zhu*, Menghui~Zhu*, Renting~Rui, Rong~Shan, Congmin~Zheng, Bo~Chen, Yunjia~Xi, Jianghao~Lin, Weiwen~Liu, Ruiming~Tang, Yong~Yu, Weinan~Zhang% <-this % stops a space
% \thanks{M. Shell was with the Department
% of Electrical and Computer Engineering, Georgia Institute of Technology, Atlanta,
% GA, 30332 USA e-mail: (see http://www.michaelshell.org/contact.html).}% <-this % stops a space
\thanks{* Equal Contribution.}
\thanks{J. Zhu, R. Rui, R. Shan, C. Zheng, Y. Xi, J. Lin, W. Liu, Y. Yu, and W. Zhang are with the Shanghai Jiao Tong University. E-mail:\{gebro13, ruirenting, shanrong, despzcm, xiyunjia, chiangel, wwliu, wnzhang\}@sjtu.edu.cn, yyu@apex.sjtu.edu.cn.}% <-
\thanks{M. Zhu, B. Chen and R. Tang are with Huawei Noah's Ark Lab. E-mail:\{zhumenghui1, chenbo116, tangruiming\}@huawei.com.}% <-this % stops a space
\thanks{Weinan Zhang is the corresponding author.}
% \thanks{Manuscript received April 19, 2005; revised August 26, 2015.}
}

% note the % following the last \IEEEmembership and also \thanks - 
% these prevent an unwanted space from occurring between the last author name
% and the end of the author line. i.e., if you had this:
% 
% \author{....lastname \thanks{...} \thanks{...} }
%                     ^------------^------------^----Do not want these spaces!
%
% a space would be appended to the last name and could cause every name on that
% line to be shifted left slightly. This is one of those "LaTeX things". For
% instance, "\textbf{A} \textbf{B}" will typeset as "A B" not "AB". To get
% "AB" then you have to do: "\textbf{A}\textbf{B}"
% \thanks is no different in this regard, so shield the last } of each \thanks
% that ends a line with a % and do not let a space in before the next \thanks.
% Spaces after \IEEEmembership other than the last one are OK (and needed) as
% you are supposed to have spaces between the names. For what it is worth,
% this is a minor point as most people would not even notice if the said evil
% space somehow managed to creep in.

% The paper headers
\markboth{Journal of \LaTeX\ Class Files,~Vol.~14, No.~8, August~2015}%
{Shell \MakeLowercase{\textit{et al.}}: Bare Demo of IEEEtran.cls for IEEE Journals}
% The only time the second header will appear is for the odd numbered pages
% after the title page when using the twoside option.
% 
% *** Note that you probably will NOT want to include the author's ***
% *** name in the headers of peer review papers.                   ***
% You can use \ifCLASSOPTIONpeerreview for conditional compilation here if
% you desire.

% If you want to put a publisher's ID mark on the page you can do it like
% this:
%\IEEEpubid{0000--0000/00\$00.00~\copyright~2015 IEEE}
% Remember, if you use this you must call \IEEEpubidadjcol in the second
% column for its text to clear the IEEEpubid mark.

% use for special paper notices
%\IEEEspecialpapernotice{(Invited Paper)}

% make the title area
\maketitle

% As a general rule, do not put math, special symbols or citations
% in the abstract or keywords.
\begin{abstract}

The advent of large language models (LLMs), such as GPT, Gemini, and DeepSeek, has significantly advanced natural language processing, giving rise to sophisticated chatbots capable of diverse language-related tasks. The transition from these traditional LLM chatbots to more advanced AI agents represents a pivotal evolutionary step. However, existing evaluation frameworks often blur the distinctions between LLM chatbots and AI agents, leading to confusion among researchers selecting appropriate benchmarks.
To bridge this gap, this paper introduces a systematic analysis of current evaluation approaches, grounded in an evolutionary perspective. We provide a detailed analytical framework that clearly differentiates AI agents from LLM chatbots along five key aspects: complex environment, multi-source instructor, dynamic feedback, multi-modal perception, and advanced capability. Further, we categorize existing evaluation benchmarks based on external environments driving forces, and resulting advanced internal capabilities. For each category, we delineate relevant evaluation attributes, presented comprehensively in practical reference tables.
Finally, we synthesize current trends and outline future evaluation methodologies through four critical lenses: environment, agent, evaluator, and metrics. Our findings offer actionable guidance for researchers, facilitating the informed selection and application of benchmarks in AI agent evaluation, thus fostering continued advancement in this rapidly evolving research domain.

\end{abstract}

% Note that keywords are not normally used for peerreview papers.

% For peer review papers, you can put extra information on the cover
% page as needed:
% \ifCLASSOPTIONpeerreview
% \begin{center} \bfseries EDICS Category: 3-BBND \end{center}
% \fi
%
% For peerreview papers, this IEEEtran command inserts a page break and
% creates the second title. It will be ignored for other modes.
\IEEEpeerreviewmaketitle

\section{\textbf{Introduction}}
\IEEEPARstart{T}he advent of Transformer~\cite{vaswani2017attention} has revolutionized natural language processing (NLP) and enabled large language models (LLMs) for chatbots, such as GPT~\cite{openai2023gpt4}, LLaMA~\cite{touvron2023llama}, Gemini~\cite{anil2023gemini}, Qwen~\cite{bai2023qwen}, and DeepSeek~\cite{guo2025deepseek}, achieving unprecedented performance across diverse text-based tasks. These models, trained on massive corpora, exhibit emergent capabilities in text generation, comprehension, and reasoning. Their ability to generalize across domains has positioned LLM chatbots as the foundation for modern AI systems, ranging from conversational interfaces to knowledge-intensive problem-solving.

\begin{figure}[t]
  \centering
  \vspace{-10pt}
  \includegraphics[width=0.48\textwidth]{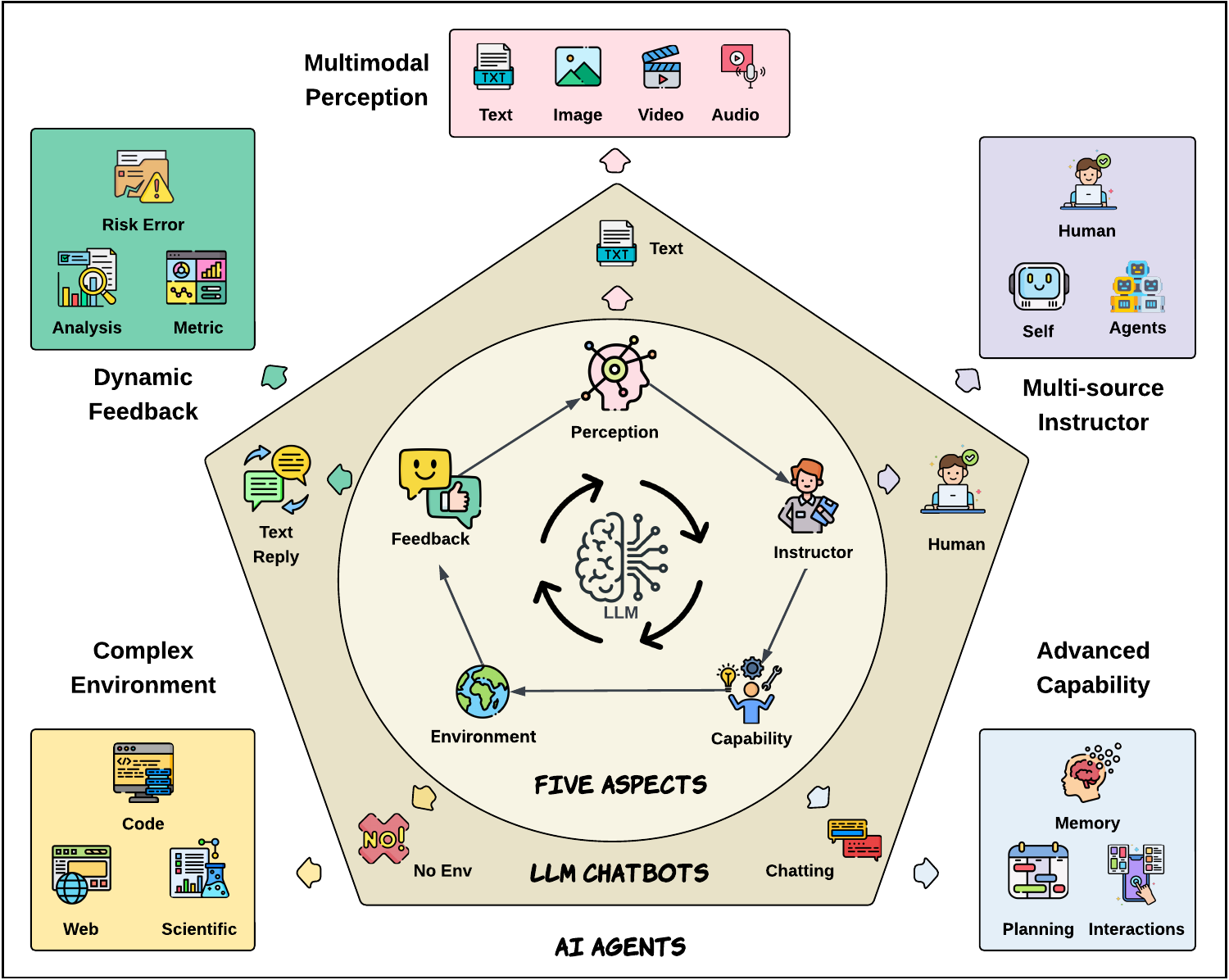}
  % \vspace{-10pt}
  \caption{Evolution between AI agents and LLM chatbots in five key aspects.}
  \label{fig: introduction}
  \vspace{-10pt}
\end{figure}

The emerging AI agents mark a further significant evolution beyond traditional LLM chatbots by enabling rich environmental interaction and broader functionality. Unlike chatbots, which only respond to human prompts in isolation, AI agents can interact with the web~\cite{Agent-E}, invoke APIs~\cite{song2024beyond}, and adapt based on real-world feedback, allowing them to handle more complex tasks~\cite{shinn2023reflexion}. 
Essentially, the transition from LLM chatbots to fully functional AI agents is an evolutionary process. Therefore, it is imperative to first clearly delineate the advancements that have occurred during this evolution.  As shown in Figure~\ref{fig: introduction}, LLM chatbots operate as reactive conversational engines, isolated from their surroundings and dependent solely on human input. In contrast, AI agents can be systematically delineated across five primary dimensions: complex environments, multi-source instructors, dynamic feedback, multi-modal perception and advanced capabilities. Details for the evolution from LLM chatbots to AI agents are illustrated in Section~\ref{sec:preliminary}.

Such a huge evolution of AI agents necessitates new philosophical and methodological approaches to AI agent evaluation, which raises our core question in this paper: 

\textbf{\textit{Given the rapid advancement of AI agents, how can we systematically evaluate LLM-based AI agents from an evolutionary perspective?}}

In response to this question, numerous evaluation benchmarks for AI agents have emerged, alongside a growing number of surveys dedicated to LLM evaluation~\cite{chang2023surveyevaluationlargelanguage,laskar2024systematicsurveycriticalreview,guo2023evaluatinglargelanguagemodels}. However, these works tend to focus either solely on LLM chatbots or provide only broad overviews of AI agent benchmarks without clearly defining the key distinctions between LLM chatbots and AI agents. Furthermore, they often overlook the evolutionary process from chatbot to agent, as well as the nuanced differences among evaluation benchmarks. As a result, researchers still face substantial ambiguity when selecting appropriate benchmarks to evaluate specific types of agents amidst a rapidly expanding landscape of evaluation tools.

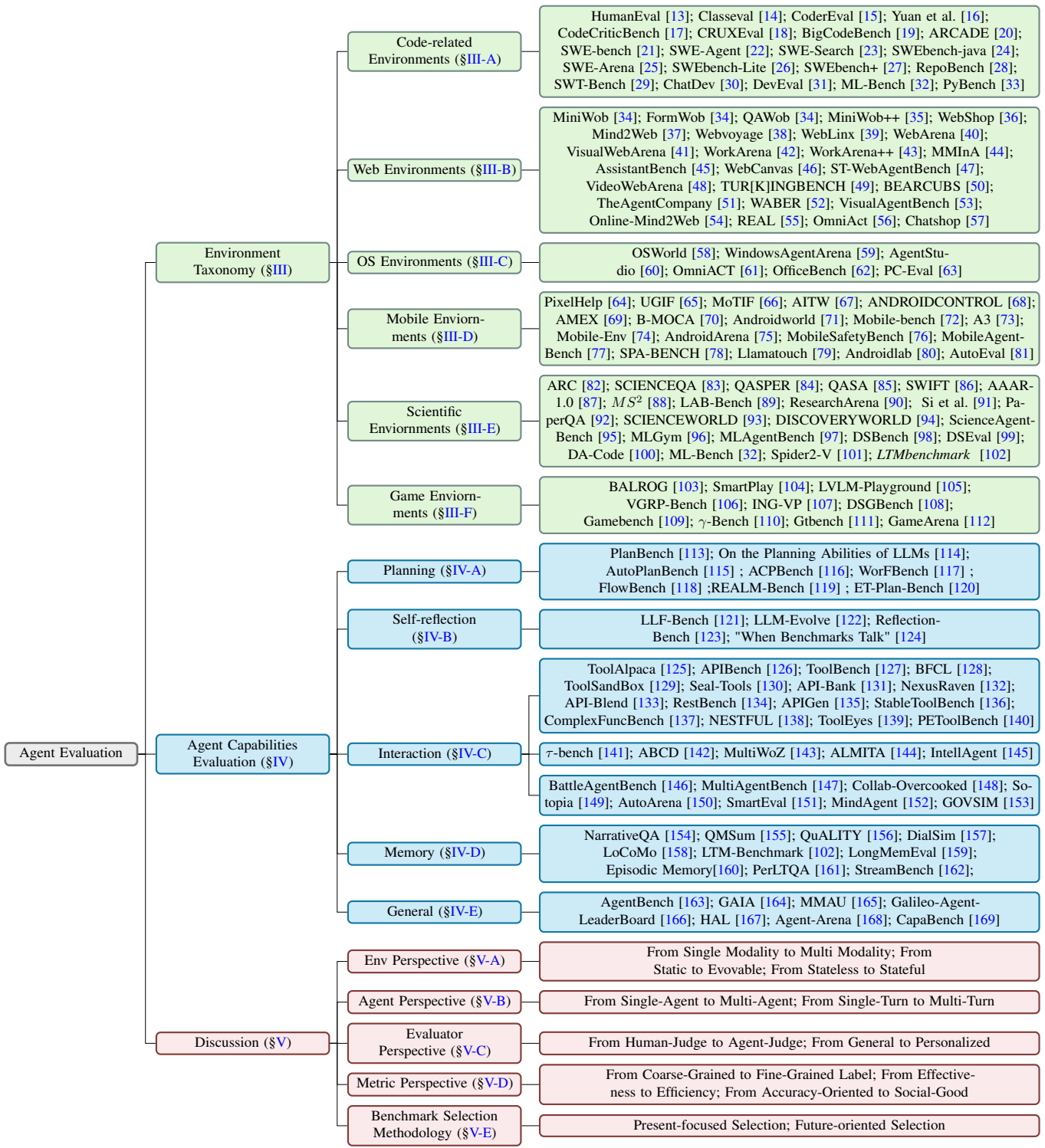
\begin{figure*}[!htb]
    \centering
    \resizebox{1\textwidth}{!}{
    \begin{forest}
        for tree={
            forked edges,
            grow'=0,
            draw,
            rounded corners,
            node options={align=center},
            text width=4cm,
            s sep=6pt,
            calign=child edge,
            calign child=(n_children()+1)/2,
            l sep=12pt,
        },
        [Agent Evaluation, root,
            [Environment Taxonomy (\S\ref{sec:environments}), env_section,
                [Code-related Environments (\S\ref{sec:code}), env_section 
                [% Use env_section for subsections too
                     HumanEval~\cite{chen2021evaluating}; 
                     Classeval~\cite{du2023classeval};
                     CoderEval~\cite{yu2024codereval};
                     Yuan et al.~\cite{yuan2023evaluating};
                     CodeCriticBench~\cite{zhang2025codecriticbench};
                     CRUXEval~\cite{gu2024cruxeval};
                     BigCodeBench~\cite{zhuo2024bigcodebench};
                     ARCADE~\cite{yin2022natural};
                    SWE-bench~\cite{jimenez2023swe};
                    SWE-Agent~\cite{yang2024swe};
                    SWE-Search~\cite{antoniades2024swe};
                    SWEbench-java~\cite{zan2408swe};
                    SWE-Arena~\cite{swe-arena2024};
                    SWEbench-Lite~\cite{yadavally2025large};
                    SWEbench+~\cite{aleithan2024swe};
                    RepoBench~\cite{liu2023repobench};
                    SWT-Bench~\cite{mundler2024swt};
                    ChatDev~\cite{qian2023chatdev};
                    DevEval~\cite{li2024prompting};
                    ML-Bench~\cite{tang2023ml};
                    PyBench~\cite{zhang2024pybenchevaluatingllmagent}
                    ,env_section, text width=12cm
                    ] % Add papers/benchmarks here
                ]
                [Web Environments (\S\ref{sec:web}), env_section
                    [{
                    MiniWob~\cite{miniwob};
                    FormWob~\cite{miniwob};
                    QAWob~\cite{miniwob};
                    MiniWob++~\cite{miniwob++};
                    WebShop~\cite{webshop};
                    Mind2Web~\cite{mind2web};
                    Webvoyage~\cite{he2024webvoyager};
                    WebLinx~\cite{lu2024weblinx};
                    WebArena~\cite{webarena};
                    VisualWebArena~\cite{visualwebarena};
                    WorkArena~\cite{workarena};
                    WorkArena++~\cite{workarena++};
                    MMInA~\cite{MMInA};
                    AssistantBench~\cite{assistantbench};
                    WebCanvas~\cite{webcanvas};
                    ST-WebAgentBench~\cite{st-webagentbench};
                    VideoWebArena~\cite{videowebarena};
                    TUR[K]INGBENCH~\cite{xu2025turkingbenchchallengebenchmarkweb};
                    BEARCUBS~\cite{bearcubs};
                    TheAgentCompany~\cite{xu2024theagentcompanybenchmarkingllmagents};
                    WABER~\cite{kara2025waber};
                    VisualAgentBench~\cite{liu2024visualagentbenchlargemultimodalmodels};
                    Online-Mind2Web~\cite{online-mind2web};
                    REAL~\cite{REAL};
                    OmniAct~\cite{OmniACT};
                    Chatshop~\cite{chatshop}
                    },env_section, text width=12cm] % Add papers/benchmarks here
                ]
                [OS Environments (\S\ref{sec:OS}), env_section
                    [
                    % \textit{LLF-Bench} \citep{cheng2023llfbenchbenchmarkinteractivelearning};
                    % \textit{LLM-Evolve} \citep{you-etal-2024-llm};
                    OSWorld~\cite{xie2024osworld};
WindowsAgentArena~\cite{bonatti2024windows};
AgentStudio~\cite{zheng2024agentstudio};
OmniACT~\cite{kapoor2024omniact};
OfficeBench~\cite{wang2024officebench};
PC-Eval~\cite{liu2025pc},
                    env_section, text width=12cm
                    ] 
                ]
                [Mobile Enviornments (\S\ref{sec:mobile}), env_section
                    [
                    PixelHelp~\cite{li2020mapping};
                    UGIF~\cite{venkatesh2022ugif};
                    MoTIF~\cite{burns2021mobile};
                    AITW~\cite{rawles2023androidinthewild};
                    ANDROIDCONTROL~\cite{li2024effects};
                    AMEX~\cite{chai2024amex};
                    B-MOCA~\cite{lee2024benchmarking};
                    Androidworld~\cite{rawles2024androidworld};
                    Mobile-bench~\cite{deng2024mobile};
                    A3~\cite{chai2025a3};
                    Mobile-Env~\cite{zhang2023mobile};
                    AndroidArena~\cite{xing2024understanding};
                    MobileSafetyBench~\cite{lee2024mobilesafetybench};
                    MobileAgentBench~\cite{wang2024mobileagentbench};
                    SPA-BENCH~\cite{chen2024spa};
                    Llamatouch~\cite{zhang2024llamatouch};
                    Androidlab~\cite{xu2024androidlab};
                    AutoEval~\cite{sun2025autoeval},
                    env_section, text width=12cm
                    ]
                ]
                [Scientific Enviornments (\S\ref{sec:scientific}), env_section
                    [
                    ARC~\cite{clark2018think};
                    SCIENCEQA~\cite{lu2022learn};
                    QASPER~\cite{dasigi2021dataset};
                    QASA~\cite{lee2023qasa};
                    SWIFT~\cite{chamoun2024automated};
                    AAAR-1.0~\cite{lou2024aaar};
                    $MS^2$~\cite{deyoung2021ms2};
                    LAB-Bench~\cite{laurent2024lab};
                    ResearchArena~\cite{kang2024researcharena};
                    ~\citet{si2024can};
                    PaperQA~\cite{lala2023paperqa};
                    SCIENCEWORLD~\cite{wang2022scienceworld};
                    DISCOVERYWORLD~\cite{jansen2406discoveryworld};
                    ScienceAgentBench~\cite{chen2024scienceagentbench};
                    MLGym~\cite{nathani2025mlgym};
                    MLAgentBench~\cite{huang2310mlagentbench};
                    DSBench~\cite{jing2024dsbench};
                    DSEval~\cite{zhang2024benchmarking};
                    DA-Code~\cite{huang2024code};
                    ML-Bench~\cite{tang2023ml};
                    Spider2-V~\cite{cao2024spider2};
                    \textit{LTMbenchmark}
                    ~\citep{castillo2024beyond},
                    env_section, text width=12cm
                    ]
                ]
                [Game Enviornments (\S\ref{sec:game}), env_section
                    [
                    BALROG~\cite{paglieri2024balrog};
                    SmartPlay~\cite{wu2023smartplay};
                    LVLM-Playground~\cite{wang2025large};
                    VGRP-Bench~\cite{ren2025vgrp};
                    ING-VP~\cite{zhang2024ing};
                    DSGBench~\cite{tang2025dsgbench};
                    Gamebench~\cite{costarelli2024gamebench};
                    $\gamma$-Bench~\cite{huang2025competing};
                    Gtbench~\cite{duan2024gtbench};
                    GameArena~\cite{hu2024gamearena},
                    env_section, text width=12cm
                    ]
                ]
            ][Agent Capabilities Evaluation (\S\ref{sec:capabilities}), capability_section [Planning (\S\ref{sec:planning}), capability_section % Use model_section for subsections too
                    [
                    PlanBench~\cite{valmeekam2023planbench}; On the Planning Abilities of LLMs~\cite{valmeekam2023planning}; AutoPlanBench~\cite{stein2023autoplanbench} ; ACPBench~\cite{kokel2025acpbench}; WorFBench~\cite{qiao2024benchmarking} ; FlowBench~\cite{xiao2024flowbench} ;REALM‐Bench~\cite{geng2025realm} ; ET‐Plan‐Bench~\cite{zhang2024plan}
                    ,
                    capability_section, text width=12cm] % Add papers/benchmarks here
                ]
                 [Self-reflection \\ (\S\ref{sec:self-reflection}), capability_section
                    [
                    LLF-Bench~\cite{cheng2023llfbenchbenchmarkinteractivelearning};
                    LLM-Evolve~\cite{llm-evolve};
                    Reflection-Bench~\cite{li2024reflectionbenchprobingaiintelligence};
                    "When Benchmarks Talk" \citep{pan2025benchmarkstalkreevaluatingcode},
                    capability_section, text width=12cm] % Add papers/benchmarks here
                ]
                [Interaction (\S\ref{sec:interaction}), 
                    capability_section
                    [
                    ToolAlpaca~\cite{toolalpaca};
                    APIBench~\cite{apibench};
                    ToolBench~\cite{toolbench};
                    BFCL~\cite{BFCL};
                    ToolSandBox~\cite{toolsandbox};
                    Seal-Tools~\cite{sealtools};
                    API-Bank~\cite{apibank};
                    NexusRaven~\cite{nexusraven};
                    API-Blend~\cite{apiblend};
                    RestBench~\cite{restbench};
                    APIGen~\cite{apigen};
                    StableToolBench~\cite{stabletoolbench};
                    ComplexFuncBench~\cite{complexfuncbench};
                    NESTFUL~\cite{nestful};
                    ToolEyes~\cite{tooleyes};
                    PEToolBench~\cite{petoolbench},
                    capability_section, text width=12cm]
                    [$\tau$-bench~\cite{yao2024tau};
                    ABCD~\cite{abcd};
                    MultiWoZ~\cite{budzianowski2020multiwozlargescalemultidomain};
                    ALMITA~\cite{ALMITA};
                    IntellAgent~\cite{yao2024intellagent},
                    capability_section, text width=12cm][BattleAgentBench~\cite{Wang2024BattleAgentBench};
                    MultiAgentBench~\cite{Gangrade2025MultiAgentBench};
                    Collab-Overcooked~\cite{Wang2025CollabOvercooked};
                    Sotopia~\cite{Wang2023SOTOPIA};
                    AutoArena~\cite{Zhao2024AutoArena};
                    SmartEval~\cite{Alahi2023SmartEval};
                    MindAgent~\cite{Yu2024MindAgent};
                    GOVSIM~\cite{Piatti2024GOVSIM},
                    capability_section, text width=12cm]
                ]
                [Memory (\S\ref{sec:memory}), 
                capability_section [NarrativeQA~\cite{kovcisky2018narrativeqa};
                    QMSum~\cite{zhong2021qmsum};
                    QuALITY~\cite{pang2021quality};
                    DialSim~\cite{kim2024dialsim};
                    LoCoMo~\cite{maharana2024evaluating};
                    LTM-Benchmark~\cite{castillo2024beyond};
                    LongMemEval~\cite{wu2024longmemeval};
                    Episodic Memory\cite{huet2025episodic};
                    PerLTQA~\cite{du2024perltqa};
                    StreamBench~\cite{wu2024streambench};
                    , capability_section, text width=12cm] % Add papers/benchmarks here
                ]
                [General
            (\S\ref{sec:general agent}), capability_section
                [ 
                AgentBench~\cite{agentbench}; GAIA~\cite{mialon2023gaiabenchmarkgeneralai}; MMAU~\cite{mmau2024};
                Galileo-Agent-LeaderBoard~\cite{galileo_agent_leaderboard2023};
                HAL~\cite{hal};
                Agent-Arena~\cite{agentarena2025};
                CapaBench~\cite{whos_the_mvp2025},
                capability_section, text width=12cm]
                ]
            ][Discussion
            (\S\ref{sec:discussion}), discussion_section
                [Env Perspective (\S\ref{sec:env perspective}), discussion_section
                [
                From Single Modality to Multi Modality; From Static to Evovable; From Stateless to Stateful,
                discussion_section, text width=12cm]
                ]
                [Agent Perspective (\S\ref{sec:agent perspective}), discussion_section
                [
                From Single-Agent to Multi-Agent; From Single-Turn to Multi-Turn,
                discussion_section, text width=12cm]
                ]
                [Evaluator Perspective (\S\ref{sec:evaluator perspective}), discussion_section
                [From Human-Judge to  Agent-Judge; From General to Personalized,
                discussion_section, text width=12cm]
                ]
                [Metric Perspective (\S\ref{sec:metric perspective}), discussion_section
                [
                From Coarse-Grained to Fine-Grained Label; From Effectiveness to Efficiency; From Accuracy-Oriented to Social-Good,
                discussion_section, text width=12cm]
                ]
                [Benchmark Selection Methodology (\S\ref{sec:benchmark_selection_methodology}), discussion_section
                [
                Present-focused Selection; Future-oriented Selection,
                discussion_section, text width=12cm]
                ]]
        ]
    \end{forest}
    }
    \caption{The overall structure of this paper. } % Caption is now in the forest label
    \label{fig:agent-evaluation-typology}
\end{figure*}

To address these gaps, we adopt an evolutionary perspective, illustrating that the development of AI agents is fundamentally driven by the interplay between \textbf{external environmental driving forces} and the resulting advanced \textbf{internal capabilities}. Accordingly, as shown in Figure~\ref{fig:agent-evaluation-typology}, we systematically categorize and synthesize the landscape of existing AI agent evaluation work using the dual lenses of Environment and Capability—the two most critical dimensions in the agent evolution process. 

% Our analysis deliberately centers on evaluation benchmarks designed for AI agents, rather than those limited to traditional LLM chatbots.\ljh{seems we also include many llm evals in capability section}

For the \textit{external environments} where agents operate, we identify several key categories in Section~\ref{sec:environments}: coding environments, web environments, operating system (OS) environments, mobile environments, scientific environments, and game environments. 
In terms of \textit{internal capabilities} in Section~\ref{sec:capabilities}, we identify planning, memory, self-reflection, and interaction as core competencies, and further discuss the current landscape of general capability evaluation. 
Moreover, for each environment and capability, we summarize a set of valuable attributes for evaluation. 
These attributes form the basis for detailed tables provided in Appendix~\ref{sec:app}, which serve as practical references for future researchers selecting benchmarks for agent evaluation.

Furthermore, Section~\ref{sec:discussion} offers a synthesis and future outlook on the evolving trends in evaluation methodology from four distinct perspectives to address a central question facing the research community: \textit{When a new agent is developed, how should the appropriate evaluation benchmark be chosen?}
We explore this question from both a present-oriented and a forward-looking perspective, aiming to provide actionable guidance for the continued advancement of agent evaluation research.

% \jc{Contribution}
The main contributions are summarized as follows:

\begin{itemize}

    \item We propose an analytical framework to distinguish AI agents from LLM chatbots from five key aspects (\ie, environment, instructor, feedback, perception, and capability), systematically characterizing the evolutionary progression from simple chatbots to advanced AI agent systems.

    \item We categorize existing evaluation benchmarks for AI agents along two critical axes: external environments and internal capabilities. For each category, detailed attribute tables and disucssions are provided to furhter benefit the research community

    \item We analyze the evolutionary trends in evaluation benchmarks across four aspects: environment, agent, evaluator, and metric. Based on these trends, we discuss future directions and offer practical guidelines on selecting appropriate benchmarks, supporting continued progress in agent evaluation research.
\end{itemize}

% \begin{figure*}[t!]
% 	\centering
% 	\resizebox{\textwidth}{!}{
% 	\input{figs/fig-tree.tex}
% 	}
% 	\caption{Structure of this paper.}
% 	\label{fig-main}
% \end{figure*}

% \definecolor{paired-light-blue}{RGB}{198, 219, 239}
% \definecolor{paired-dark-blue}{RGB}{49, 130, 188}

\definecolor{paired-light-blue}{RGB}{193, 229, 245}
\definecolor{paired-dark-blue}{RGB}{17, 121, 162}

\definecolor{paired-light-orange}{RGB}{251, 208, 162}
\definecolor{paired-dark-orange}{RGB}{230, 85, 12}
% \definecolor{paired-light-green}{RGB}{199, 233, 193}
% \definecolor{paired-dark-green}{RGB}{49, 163, 83}
\definecolor{paired-light-green}{RGB}{217, 242, 208}
\definecolor{paired-dark-green}{RGB}{114, 139, 134}
\definecolor{paired-light-purple}{RGB}{218, 218, 235}
\definecolor{paired-dark-purple}{RGB}{117, 107, 176}
\definecolor{paired-light-gray}{RGB}{217, 217, 217}
\definecolor{paired-dark-gray}{RGB}{99, 99, 99}
\definecolor{paired-light-pink}{RGB}{222, 158, 214}
\definecolor{paired-dark-pink}{RGB}{123, 65, 115}
\definecolor{paired-light-red}{RGB}{231, 150, 156}
\definecolor{paired-dark-red}{RGB}{131, 60, 56}
\definecolor{paired-light-yellow}{RGB}{231, 204, 149}
\definecolor{paired-dark-yellow}{RGB}{141, 109, 49}
\definecolor{light-green}{RGB}{118, 207, 180}
\definecolor{raspberry}{RGB}{228, 24, 99}

\tikzset{%
    root/.style =          {align=center,text width=3cm,rounded corners=3pt, line width=0.5mm, fill=paired-light-gray!50,draw=paired-dark-gray!90},
    % root/.style =          {align=center,text width=3cm,rounded corners=3pt, line width=0.5mm, fill=paired-light-gray!50,draw=paired-dark-gray!90, yshift=10cm}, % Added yshift here
    env_section/.style =  {align=center,text width=4cm,rounded corners=3pt, fill=paired-light-green!80,draw=paired-dark-green!100,line width=0.4mm},
    capability_section/.style = {align=center,text width=4cm,rounded corners=3pt, fill=paired-light-blue!80,draw=paired-dark-blue!100,line width=0.4mm},
    training_section/.style = {align=center,text width=4cm,rounded corners=3pt, fill=paired-light-green!50,draw=paired-dark-green!80, line width=0.4mm},
    inference_section/.style = {align=center,text width=4cm,rounded corners=3pt, fill=paired-light-red!35,draw=paired-light-red!90, line width=0.4mm},
    protocol_section/.style = {align=center,text width=4cm,rounded corners=3pt, fill=paired-light-yellow!40,draw=paired-dark-yellow!100, line width=0.4mm},
    discussion_section/.style = {align=center,text width=4cm,rounded corners=3pt, fill=paired-light-red!20,draw=paired-dark-red!100, line width=0.4mm},
    general_section/.style = {align=center,text width=4cm,rounded corners=3pt, fill=paired-light-purple!35,draw=paired-dark-purple!90, line width=0.4mm},% Using light red for Inference as in the example
    subsection/.style =    {align=center,text width=3.5cm,rounded corners=3pt}, % General subsection style, NO COLOR DEFINITION NOW
}

\section{\textbf{From LLM Chatbots to AI Agents: Preliminary}}
\label{sec:preliminary}

In this section, we discuss the evolutionary progression from LLM chatbots to AI agents from five key aspects, which leads to the taxonomy on agent evaluation benchmarks in later sections.
% Building on the causal relationships that drive this evolution, we further introduce our taxonomy, which systematically categorizes these developments.

\subsection{\textbf{Background: Large Language Model (LLM) Chatbots}}

The advent of Transformer architectures has laid the foundation for the current era of large language models (LLMs).
% The advent of Transformer architectures has laid the foundation for the current era of large language models (LLMs) , exemplified by breakthroughs such as GPT~\cite{openai2023gpt4}, LLaMA~\cite{touvron2023llama}, Gemini~\cite{anil2023gemini}, Qwen~\cite{bai2023qwen}, and DeepSeek~\cite{guo2025deepseek}. 
These models, trained on vast and diverse textual corpora, have demonstrated remarkable performance in language understanding~\cite{openai2023gpt4}, generation~\cite{anthropic2025claude}, and reasoning~\cite{guo2025deepseek}. Within this landscape, the LLM chatbot paradigm has emerged as a dominant application. LLM chatbots are designed as reactive conversational systems that receive textual prompts from users, process these inputs using pretrained knowledge, and generate coherent textual responses. Typical use cases include open-domain dialogue, customer support, question answering~\cite{guo2025deepseek}. 
Despite their versatility, traditional LLM chatbots are limited in that they operate in closed environments, devoid of active perception or real-world context. Their interactions are fundamentally constrained to static, turn-based exchanges, lacking the means to sense or affect their operational environment.

\subsection{\textbf{The Emergence and Definition of AI Agents}}

Recent advances have shifted the focus from reactive LLM chatbots to proactive AI agents. 
AI agents can actively interact with the environment and make decisions by processing language, perceiving, reasoning about multi-modal inputs, and acting within dynamic contexts.
Representative examples include web search agents~\cite{Agent-E}, API-calling agents for online tasks~\cite{song2024beyond}, and code agents that iteratively debug based on execution feedback~\cite{shinn2023reflexion}. 
Figure~\ref{fig: introduction} summarizes five key distinctions—perception, instructor, capability, environment, and feedback—showcasing how AI agents advance far beyond traditional chatbots in scope and intelligence.

\textbf{Complex Environment}.
The most fundamental distinction lies in the environment dimension, which serves as the primary external driving force behind agent evolution. Traditional LLM chatbots are confined to closed environments, interacting solely with humans through static dialogue and lacking awareness or control over their surroundings. In contrast, AI agents operate within diverse and complex environments, \eg, software platforms, scientific computing, Internet ecosystems and operating systems. 
This enables them to interpret dynamic contexts and take actions that affect the external world, transforming them from passive responders into proactive collaborators and task executors.

\textbf{Multi-source Instructor}.
AI agents also advance the instructor dimension. Unlike LLM chatbots that depend heavily on human prompts, agents integrate instructions from multiple sources, including self-reflection, collaboration with other agents, and hierarchical commands in multi-agent systems. This multi-source guidance empowers agents to make complex decisions, self-correct, resulting in higher autonomy and robustness.

\textbf{Dynamic Feedback}.
While LLM chatbots primarily receive feedback through conversation or user correction, AI agents operate in environments with continuous, multifaceted feedback—including metric-based analyses, risk assessments, explicit error signals, and environment-derived rewards or penalties. This rich feedback ecosystem enables ongoing adaptation, self-improvement, and long-term optimization.

\textbf{Multimodal Perception}.
% Moreover, AI agents advance significantly in Perception. 
To function in real-world settings and respond to complex instructions, AI agents are equipped with multimodal sensing—processing not just text, but also visual, auditory, and even tactile or environmental sensor data. The development of Multimodal Large Language Models (MLLMs) exemplifies this leap, allowing agents to understand and reason across diverse modalities and vastly expanding their intelligence and applicability.

\textbf{Advanced Capability}.
Advancements across environment, instructor, feedback, and perception dimensions collectively drive the evolution of agents' internal capabilities. Dynamic environments, richer instructions and feedback, and enhanced perception propel AI agents far beyond basic conversations. Agents now exhibit complex planning, persistent memory, adaptive reasoning, and autonomous task execution. This marks a transformative step in intelligent systems, demonstrating how external demands and internal advances together drive the shift from reactive LLM chatbots to autonomous AI agents.

The progression from LLM chatbot to AI agent is primarily catalyzed by two forces: \textbf{external environment} complexity and the corresponding development of \textbf{internal capabilities}. 
% As the saying goes, external driving forces stimulate internal growth.
External driving forces stimulate internal growth.
As agents are deployed in richer, more demanding environments, they must evolve to incorporate sophisticated perception, planning, and adaptation mechanisms. This external-internal interplay underlies the rapid advancement of modern AI agents. 
In the next sections, as shown in Figure~\ref{fig:agent-evaluation-typology}, we categorize and discuss agent evaluation from both external environment and internal capability perspectives.

\section{\raisebox{-0.7ex}{\includegraphics[height=3ex]{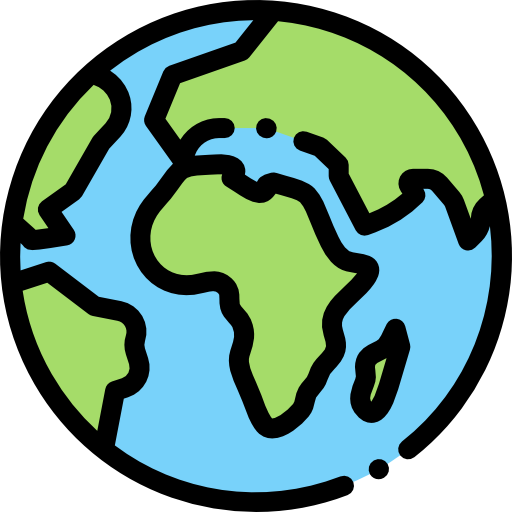}}~\textbf{Evaluation for Different Environments} }
\label{sec:environments}
% 环境在agent中扮演了一个重要的地位，环境决定了agent的交互方式，agent能完成的任务，应用场景，以及agent的设计等等
% 本章节从环境的角度来叙述benchmarks，同时介绍agent完成的一些任务，他们具体的一些区别之类的
% \mh{the definition, importance of environment; in this section, we will xxx}
The environment, as an abstraction and simplification of the world, plays a pivotal role in the evolution from LLM chatbots to AI agents. It is an external system that the agent interacts directly with. Agents are equipped with different types of environments tailored to task categories and complexity levels, enabling capabilities such as self-reflection and multi-step planning.
In this section, a systematic categorization and analysis of the environments employed in various agent evaluations is presented. 
The objective is to delineate their developmental trajectory and discern the similarities and distinctions among different environments.

\subsection{\raisebox{-0.7ex}{\includegraphics[height=3ex]{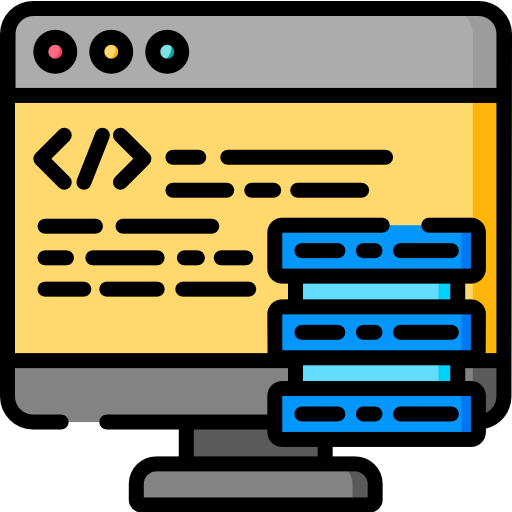}}~\textbf{Coding Environments}}
\label{sec:code}
% 常见的交互环境是xxx，涉及到的任务是哪些
In the context of general-purpose agents, codes represent a pivotal means of interactive engagement with the external environment and exerting influence on it. Consequently, the coding abilities are of paramount importance for agents. The common environments for code agents include referenced code files, code repositories, code executors and so on, which are determined by specific tasks.

% agent时代之前，不涉及交互环境，具体做哪些任务，评估的指标是哪些
% 或者说很难称得上是环境，只有instructor和model
It is worthy of mentioning that a number of the earlier works are not related to code environments, in which case the benchmarks are only evaluated on simple tasks and the agents involved are more akin to code LLMs. We will only briefly list these benchmarks and describe the simple tasks on which they were evaluated~\cite{chen2021evaluating,du2023classeval,yu2024codereval,yuan2023evaluating,zhang2025codecriticbench,gu2024cruxeval,zhuo2024bigcodebench,chen2021evaluating}.
As a pioneer of this area, HumanEval~\cite{chen2021evaluating} tests the LLMs by asking them to synthesise programs from docstrings.
Likewise, CoderEval~\cite{yu2024codereval} enhances HumanEval mainly on the complexity of the tasks, especially on the cyclomatic complexity and lines of codes.
Similarly, CodeCriticBench~\cite{zhang2025codecriticbench} assesses LLMs' aptitude for code critique, with a particular emphasis on code generation and quality assurance (QA) tasks.
BigCodeBench~\cite{zhuo2024bigcodebench} enhances HumanEval~\cite{chen2021evaluating} and offers more challenging and practical tasks, which has higher requirements for LLMs like utilizing diverse function calls as tools.
ARCADE~\cite{yin2022natural} is also an evaluation on code generation, but it pays attention to interactive notebooks in data science.

% 在agent的时代，交互环境有哪些. 然后具体介绍做了些啥
As code-related tasks become more complex and comprehensive, it is no longer enough for agents to take in only input from the instructors. They need to operate on more objects and performs skills such as self-reflection, where the environment provides the basis for the agents.

To better understand the environments in code agents, here we take the representative benchmark on code agents SWE-bench~\cite{jimenez2023swe} as the example. 
In SWE-bench, agents are prompted to understand the real issues of GitHub and generate the code patches to resolve the issues. During inference, related codes and demonstrations are inserted into the prompts beforehand and LLMs directly generate the patches in text, where the models behave more like semi-agents. The environment now only includes the codebases and the automated unit tests. In original SWE-bench paper, the best-performing model, Claude 2, though equipped with a simple retrieval mechanism, could only solve 1.96\% of the issues. After that, in order to archieve better performances on these tasks, a series of agent frameworks are proposed to challenge the tasks on SWE-bench such as SWE-Agent~\cite{yang2024swe} and SWE-Search~\cite{antoniades2024swe}. To support the agents' abilities such as memory and self-reflection, an external terminal is added into the environment, through which the agents could perform some higher-level operations like file editing and searching.
Except SWE-bench, a series of similar benchmarks are proposed and their environments differs from that in SWE-bench mainly on the modal~\cite{yang2024swe}, programming language~\cite{zan2408swe}, agents' comparison~\cite{swe-arena2024}, execution-free evaluation~\cite{yadavally2025large} and data quality~\cite{aleithan2024swe}.

In addition, in the field of code agents, there are various benchmarks with different environments towards different specific code-related tasks. RepoBench~\cite{liu2023repobench} evaluates on the tasks of code retrieval and code completion and provides the agents with the code repositories as the environment. Similarly, for the task of code completion at the repository level, the environment in RepoEval~\cite{zhang2023repocoder} provides the capability of retrieval. This is the basis of the iterative retrieval-generation mechanism in the paper. SWT-Bench~\cite{mundler2024swt} is based on SWE-bench and adopts similar environments to benchmark agents on the generation of test cases according to the issues. Based on ChatDev~\cite{qian2023chatdev}, DevEval's~\cite{li2024prompting} environments allow to evaluate agent on the full stages of software development, including software design, environment setup, implementation and testing. ML-Bench~\cite{tang2023ml} offers a ML-related repository and a Linux sandbox and ask the agents to perform machine learning tasks. Pybench~\cite{zhang2024pybenchevaluatingllmagent} introduces a Python code interpreter to the environment. The agents in the benchmark operate on the different type of files in the environment to complete some real-world tasks like chart analysis, text analysis, image \& audio editing and so on.
% discussion：从llm到agent，为啥evaluation的方式变了
% \textbf{\textit{Discussion.}}  ss

\subsection{\raisebox{-0.7ex}{\includegraphics[height=3ex]{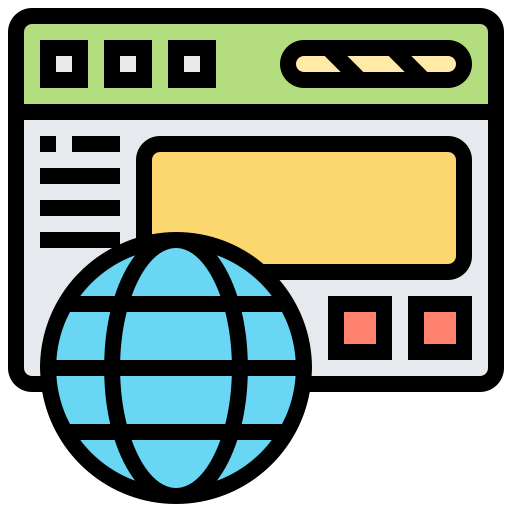}}~\textbf{Web Environments}}
\label{sec:web}

Web agents, autonomous systems that navigate, interact with, and extract information from web pages, have emerged as a powerful paradigm for tasks ranging from form filling and information retrieval to complex multi‑step workflows on enterprise platforms. Evaluating their performance is crucial for monitoring progress and identifying remaining challenges. Over the past few years numerous benchmarks have been proposed; we summarize them in Table~\ref{tab:Web Env} and organize our discussion by the \emph{realism} of the environment.                            

% \textbf{Synthetic Environments.}
Early benchmarks use fully synthetic web pages, designed to stress reinforcement‑learning agents in a controlled setting. Two pioneering works are MiniWoB~\cite{miniwob} and MiniWoB++~\cite{miniwob++}. MiniWoB provides a suite of simple tasks (e.g., clicking buttons, filling out toy mail and calendar forms) rendered in canvas elements. Its synthetic nature allows reproducible offline RL training and evaluation. MiniWoB++ extends MiniWoB with longer sequences, random layouts, and soft‑text reasoning tasks (e.g., checkbox grids, multi‑layout navigation, simple social‑media interactions), further challenging sequence modeling and exploration.
These early synthetic suites established the importance of standardized evaluation for web agents and paved the way for more realistic benchmarks.

% \textbf{Semi‑Real Environments.}
As language‑model‑based agents matured, purely synthetic data proved insufficient. Semi‑real benchmarks load snapshots of real websites and host simplified or anonymous modifications on data replicas to afford reproducible evaluation on genuine layouts and data. FormWoB~\cite{miniwob} offers form-filling tasks to assess agents' abilities in handling structured inputs. QAWoB~\cite{miniwob} designs question-answering tasks to test agents' information retrieval and comprehension skills. WebShop~\cite{webshop} simulates online shopping scenarios, challenging agents in multi-step decision-making and understanding user preferences. ChatShop~\cite{chatshop} combines conversational systems with shopping tasks, evaluating agents' consistency and task completion in multi-turn dialogues. WebArena~\cite{webarena} constructs simulated environments based on real websites, encompassing various web interaction tasks to test agents' generalization capabilities. Its variants VisualWebArena~\cite{visualwebarena} and VideoWebArena~\cite{videowebarena} incorporates visual information and video tutorials, respectively, challenging agents' performance in processing mixed text, image content and long-context videos. WABER~\cite{kara2025waber} provides a multilingual, multi-task evaluation environment to test agents' cross-lingual capabilities and task adaptability. REAL~\cite{REAL} recreates eleven high‑fidelity website simulations  with deterministic snapshots and combines script‑based state validation with an LLM‑based rubric for open‑ended retrieval tasks.
ST‑WebAgentBench~\cite{st-webagentbench} offers a framework for evaluating agents in multi-turn dialogues and complex tasks, testing long-term planning and context understanding. TUR[K]INGBENCH~\cite{xu2025turkingbenchchallengebenchmarkweb} uses original Mechanical Turk HTML pages, evaluating agents on naturally crowdsourced web tasks with metrics like ROUGE‑L and IoU, underscoring diversity of real‑world page designs. TheAgentCompany~\cite{xu2024theagentcompanybenchmarkingllmagents} simulates a small software company environment, focusing on agents' performance in real-world professional scenarios. 
% \mh{what's the feature of these environments?}

% \textbf{Fully Dynamic Real-World Web Benchmarks.}
In contrast to the aforementioned semi-realistic benchmarks, a growing number of fully dynamic and real-world benchmarks have recently been proposed. These benchmarks either encapsulate real environments into snapshot-based simulations or directly connect to the live Internet for task execution. Such evaluation paradigms better reflect real-world complexities and are more effective at assessing agents' actual performance and robustness in dynamic conditions. Mind2Web~\cite{mind2web} requires agents to understand rich page structures and plan multi-step goals. Online-Mind2Web~\cite{online-mind2web} further extends this by introducing challenges related to dynamic content changes and real-time DOM variations. WebVoyager~\cite{he2024webvoyager} emphasizes effective intermediate state modeling and reasoning over long-term dependencies between distant pages. WebLINX~\cite{lu2024weblinx} focuses on reasoning through numerous branching paths and integrating historical click sequences to reach target information.
WorkArena~\cite{workarena} and WorkArena++~\cite{workarena++} assess agents'  long-context understanding, reflecting more realistic workflows encountered by office or workers in real enterprises.
MMInA~\cite{MMInA} introduces challenges in multimodal and multi-hop web tasks.
AssistantBench~\cite{assistantbench} mainly tests agents' ability to follow time-consuming and multi-turn workflows.
WebCanvas~\cite{webcanvas} poses challenges in visual parsing and UI element grounding under dynamic layouts by examining intermediate nodes.
BEARCUBS~\cite{bearcubs} centers on knowledge-intensive tasks that require agents retrieve and utilize webpage content and external knowledge sources.

% WebArena
\subsection{\raisebox{-0.7ex}{\includegraphics[height=3ex]{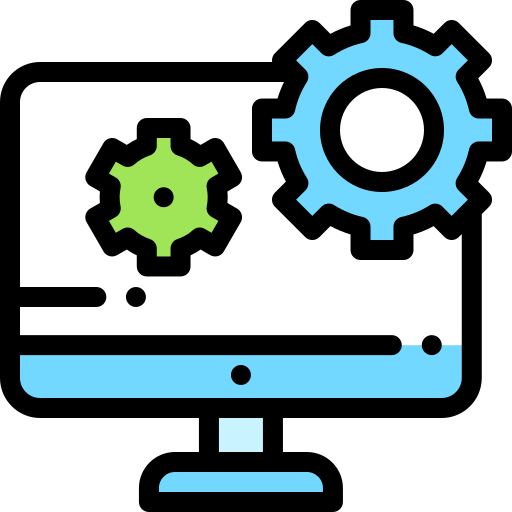}}~\textbf{OS Environments}}
\label{sec:OS}
Despite the web environments, there are some benchmarks that are based on the desktop, which is more complex and challenging. OSWorld~\cite{xie2024osworld} serves as a representative desktop benchmark, demonstrating the optimal characteristics of a desktop environment. It employs virtual machines (VMs) to facilitate a desktop environment that is executable and controllable. The agents could interact with the environment just like operating the computer in reality. The environments are initialized by config files, which describes some operations like open a file after starting the virtual machine. During the interaction process with the environment, the observation of agents encompasses the capture of screenshots of the desktop and the accessibility (a11y) tree. The action space defined in OSWorld is diverse and it supports all mouse and keyboard actions, including movement, clicks and so on. There are totally 369 evaluated tasks in OSWorld, including Office tasks, OS tasks, Daily tasks, Workflow tasks and Professional tasks.

Furthermore, there are several distinct benchmarks that emphasize disparate fine-grained scenarios. 
Different from OSWorld, whose tasks are mainly on Ubuntu, WindowsAgentArena~\cite{bonatti2024windows} offers a general environment focusing on the Windows operating system.
AgentStudio~\cite{zheng2024agentstudio} constitutes a triad of virtual agent research environments, tools, and benchmarks. AgentStudio's environment is conducive to the observation of text, image, and video formats. Its action space supports both GUI operations and API calls.
OmniACT~\cite{kapoor2024omniact} provides a static dataset for the evaluation of generalist autonomous agents. This dataset contains tasks in natural language, screenshots, and PyAutoGUI scripts that serve as ground truth.
The environment at OfficeBench~\cite{wang2024officebench} is characterized by its concentration on office tasks, which distinguishes it from the general desktop agent evaluation environments, such as OSWorld. The action space in OfficeBench is restricted and depends on the applications that agents are currently utilizing. For instance, the action of \textit{run\_command} could be operated in Shell but invalid in other applications like Word. In OfficeBench, agents could observe both the current and the previous states and actions, which is a departure from the settings in previous benchmarks.
Although analogous to OSWorld, PC-Eval~\cite{liu2025pc} augments it with more practical and challenging tasks, which are checked by human annotators.
\subsection{\raisebox{-0.7ex}{\includegraphics[height=3ex]{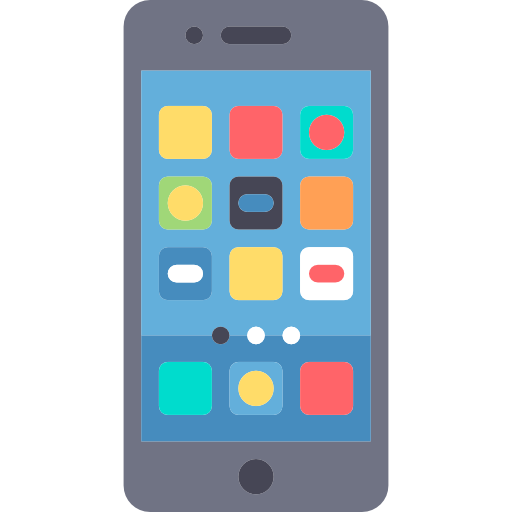}}~\textbf{Mobile Environments}}
\label{sec:mobile}
% mobile agent主要tasks envrionments里包含什么
% static环境，文本环境
% dynamic环境，真实可以调用的环境，GUI环境
% discussion 单轮to多轮 验证成功的方式/是否需要人工介入 自动化/单app到跨app

The advent of mobile Internet has led to the gradual integration of mobile devices, such as smartphones and tablet computers, into our daily life. These devices have become increasingly indispensable, playing a crucial role in various aspects of our lives. However, the increasing functionality of mobile devices poses challenges to their usability. Many mobile agents such as Siri, Bixby and XiaoAi have been proposed to relieve the operation issues. Nevertheless, in general,  there is still significant room for improvement in task completion capabilities, personalization, and other aspects for these agents. Therefore, many benchmarks are proposed to provide the testbeds and promote the development of mobile agents.

The environments of earlier benchmarks are usually static, or in other way to say, are just fixed datasets to evaluate LLM chatbots. Early works such as PixelHelp~\cite{li2020mapping}, UGIF~\cite{venkatesh2022ugif}, MoTIF~\cite{burns2021mobile} and AITW~\cite{rawles2023androidinthewild} offer the LLMs with the screenshots and tree-based representations such as View Hierarchy (VH) and regard the demonstrations from the humans as ground truths to evaluate the agents. Among these benchmarks, the process of evaluation is static and does not need the involvement of executable environments, where the LLMs could not interact with the environments in multiple rounds. Besides the ones mentioned above, ANDROIDCONTROL~\cite{li2024effects} mainly improves on the granularity of the instructions and the freshness of the APPs. AMEX~\cite{chai2024amex} further offers the screen descriptions and screen elements for the agents.

% Androidinthewild: A large-scale dataset for android device control(static dataset)
% Amex: Android multi-annotation expo dataset for mobile gui agents (static dataset)
% androidcontrol(On the effects of data scale on ui control agents) (static)

However, when the environments are asked to evaluate AI agents rather than LLM chatbots, things get a little different. The complexity of the tasks inherently determines the requirements of multi-step execution. In addition, the agents usually need multi-round interactions to handle their inner mechanisms such as planning and self-reflection. Also, there are usually multiple trajectories to complete the tasks, which makes static datasets even less suitable for evaluating mobile agents.
In the era of agents, the environments typically include an emulator and an adapter. The emulator, which is the core of the environment, behaves like a real mobile device and responds to the actions of the agents. For example, if the agents give the action of tapping and dragging up, the emulator will scroll down the screen. The adapter, on the other hand, acts as a bridge between the emulator and the agents. When the agents perform actions, the adapter translates the agents' instructions into shells and passes them to the emulator through tools such as ADB. Similarly, it gets the states of the environment such as the screenshot and XML from the emulator and converts them into the format that the agents can understand.

With the above background knowledge, we now proceed to a detailed characterization of existing evaluations. 
B-MOCA~\cite{lee2024benchmarking} is an earlier benchmark with dynamic environments and includes a simple range of daily tasks like web tasks, shopping tasks, system tasks and so on.
Androidworld~\cite{rawles2024androidworld} expands the scope of APPs. It's equipped with the ability to dynamically construct the initial states of tasks and vary the task parameters in unlimited ways.
Mobile-bench~\cite{deng2024mobile} extends the agent's action space with API operations, allowing the agent to perform tasks such as opening a specific APP without screen operations. Additionally, to evaluate the agents more accurately, it proposes checkpoints to evaluate the agents on whether reach essential points during the planning and reasoning steps. 
A3~\cite{chai2025a3} focuses on the daily used APPs and offers a more flexible action space for the agents, which supports additional actions like Long Press.
The environments in Mobile-Env~\cite{zhang2023mobile} supports intermediate rewards and intermediate instructions for the agents.
AndroidArena~\cite{xing2024understanding} pays more attention to the cross-APP tasks and constrained tasks. The environment of cross-APP tasks involves multiple APPs and the agents require cooperation between multiple APPs. Besides for the regular instructions, the environment should also offer constraints such as \textit{"do not click the payment button"}. MobileSafetyBench~\cite{lee2024mobilesafetybench} develops a set of safety-related tasks to evaluate the mobile agents, such as messaging and banking applications.

In addition to the aforementioned benchmarks, there are works that mainly improve the evaluation methods. This is also a key point of the environments. 
MobileAgentBench~\cite{wang2024mobileagentbench} utilizes Android Accessibility Service to capture app events and uses them to check whether the task is completed.
SPA-BENCH~\cite{chen2024spa} proposes a coarse-to-fine success detection pipeline to judge if an agent succeeds. It first applies coarse detection with Optical Character Recognition (OCR) and matches the trajectories with the key components. After that, it deploys a trained multimodal LLM for final success determination.
Llamatouch~\cite{zhang2024llamatouch} automatically evaluates agents by whether the agent traverses all manually annotated application / system states.
Similarly, in Androidlab~\cite{xu2024androidlab}, each task is divided into multiple page states as sub-goals. A task is considered complete only if all sub-goals are correctly addressed.
AutoEval~\cite{sun2025autoeval} adopts an automatic judge system to evalute the agents, which composed of three components: Capturer, Reasoner and Checker. The Capturer generates descriptions about the screenshots and the Reasoner judge the agent's performance according to the screen and task descriptions. Finally, the Checker is added to guarantee the Reasoner’s output is acceptable and consistent with the reasoning process.

\subsection{\raisebox{-0.7ex}{\includegraphics[height=3ex]{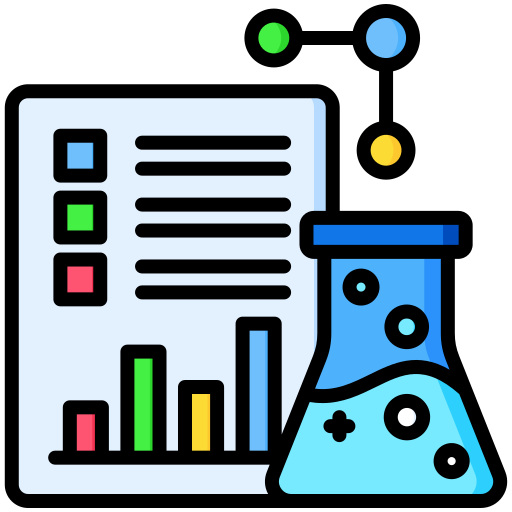}}~\textbf{Scientific Environments}}
\label{sec:scientific}
The advancements of AI agents have sparked interest in building the entire automated process of scientific discovery or assist researchers in different researching stages. Early benchmarks mainly assess the scientific reasoning and knowledge retrieval capabilities of LLM chatbots. For example, ARC~\cite{clark2018think} and SCIENCEQA~\cite{lu2022learn} evaluate LLM chatbots on scientific multi-choice question answering. QASPER~\cite{dasigi2021dataset} and QASA~\cite{lee2023qasa} focus on information seeking and answer anchoring in research papers. SWIFT~\cite{chamoun2024automated} measures the research paper comment generation and weakness identification abilities. AAAR-1.0~\cite{lou2024aaar} evaluates LLM chatbots on research tasks of equation inference, experiment design, paper weakness, and review critique. $MS^2$~\cite{deyoung2021ms2} emphasizes multi-document summarization of medical studies, while LAB-Bench~\cite{laurent2024lab} measures a broad range of tasks for scientific research, \textit{e.g.} analysis of tables and figures. These benchmarks are typically conducted in an offline static manner, and there is no clear involvement of an external environment. 

Instead of a straightforward evaluation, there are also benchmarks involving a scientific article pool as an external environment, which is utilized for retrieval and assists in different research stages, including academic surveying, idea generation, experiment design, writing assistance and so on. For example, ResearchArena~\cite{kang2024researcharena} benchmarks agents' ability to collect surveys and organize information. \citet{si2024can} compares the ideas generated by research agents with expert NLP researchers, which are based on paper retrieval for RAG. PaperQA~\cite{lala2023paperqa} performs information retrieval across full-text scientific articles for question answering. These agents harness extracted knowledge from the environment, but lack explicit interaction.

Since developing and evaluating an agent's capacity for real-world scientific discovery is often expensive and challenging, some benchmarks construct a simulated, interactive and text-based environment to evaluate the agents' capabilities. Representatives of them are SCIENCEWORLD~\cite{wang2022scienceworld} and DISCOVERYWORLD~\cite{jansen2406discoveryworld}. SCIENCEWORLD tests agents' scientific reasoning capabilities in a text environment with a simulation for thermodynamics, electrical circuits, chemical reactions, and biological processes. DISCOVERYWORLD benchmarks agents' ability to perform complete cycles of scientific discovery.

As the applications of scientific agents become more complex, the demands on their capabilities continue to grow, and the environments in which they interact are also becoming more complex and realistic. Specifically, in order to automate the complete cycle of scientific discovery, agents may need to interact with the environment by reading in-memory data, handling file permission, editing files, executing scripts, and collecting results. In this case, the environment is complex, which can be composed of file systems, code interpreter, shell, database and so on. We refer to this compound environment as \textit{workspace}, which is typically a jupyter notebook or a virtual machine in many cases. ScienceAgentBench~\cite{chen2024scienceagentbench} evaluates agents in an environment equipped with bash shell, web browser, code interpreter on coding for workflow of scientific discovery. MLGym~\cite{nathani2025mlgym} designs a gym-like environment where agents can manipulate with shell, tools, python dependencies and permission for various files. MLAgentBench~\cite{huang2310mlagentbench}, DSBench~\cite{jing2024dsbench} and MLE-bench~\cite{chan2024mle} mainly focus on Kaggle challenges and completions, assessing agents' end-to-end machine learning engineering capabilities. The workspace allows file editing and code running by these research agents. DSEval~\cite{zhang2024benchmarking} also assesses the performance of agents throughout
the entire data science lifecycle, with a runtime session (\textit{i.e.} jupyter notebook) composed of in-memory data, external files, execution history and code executor. DA-Code~\cite{huang2024code} and ML-Bench~\cite{tang2023ml} evaluate agents in a Docker on data science code generation and repository-level code manipulation respectively. Spider2-V~\cite{cao2024spider2} focuses on multimodal agents for data science automation, which also relies on a Docker environment  for file transfer, application launch, remote API calls, script execution and playright automation.

\subsection{\raisebox{-0.7ex}{\includegraphics[height=3ex]{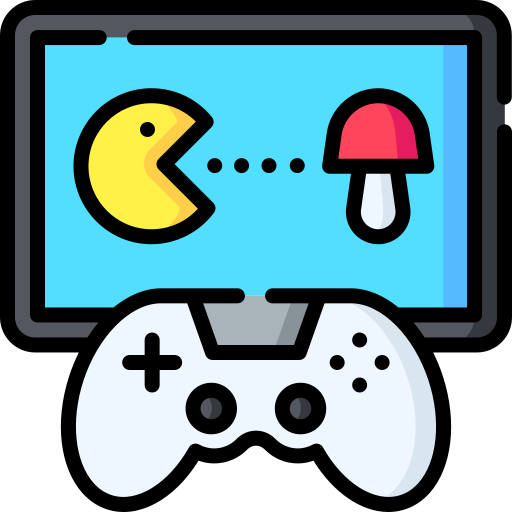}}~\textbf{Game Environments}}
\label{sec:game}
Prior to the advent of large language models, game environments were extensively utilized as testbeds for AI agents. In contrast to agents such as code agents and web agents, the value in application of agents in game environments is relatively low. In general, benchmarks with game environments do not evaluate agents on specific tasks; rather, they aim to evaluate general capabilities such as planning and reasoning.
% Furthermore, the distinction between evaluation on LLM chatbots and AI agents in game environments remains ambiguous. This phenomenon can be attributed to the interactive nature of the game environment, which prompts LLMs to emulate the behaviors of agents. 
In this section, we primarily introduce several game environments that have been utilized in existed benchmarks for agents.

One branch of game environments for agent evaluations involve only one agent and focus on the absolute capabilities of the agents.
BALROG~\cite{paglieri2024balrog} evaluates agents on common games like BabyAI~\cite{carta2023grounding}, Crafter~\cite{hafner2021benchmarking}, TextWorld~\cite{cote2019textworld} and so on. For VLMs (Vision Language Model), the environments provide directly the screenshots of the games. While for LLMs, all of the observations in games are transformed into texts for the agents to understand.
SmartPlay~\cite{wu2023smartplay} provides a comprehensive evaluation of the agent's capabilities in various domains, including long text understanding, reasoning, instruction/rule following, and so on. The evaluation process encompasses a diverse array of both visual and text-based games, including Bandits, Rock-Paper-Scissors, Tower of Hanoi, Minecraft, Messenger, and Crafter. In certain environments, internal rewards are utilized as the evaluation metric, while in others, completion rates are employed to assess the performance of the agents.
In a similar manner, LVLM-Playground ~\cite{wang2025large} adopts the games including TicTacToe, Reversi, Sudoku to evaluate the agents' capabilities on perception, reasoning, decision and adversary. The evaluation metrics are multifaceted and differ depending on the specific task. For instance, for perceiving task, which asks the agents to convert a visual game state into a structured matrix representation, and the accuracy is utilized as the metric.
In contrast to the aforementioned work, VGRP-Bench~\cite{ren2025vgrp} focuses on puzzle games and employs them as a means to assess the agents.
ING-VP~\cite{zhang2024ing} focuses evaluation of multi-step planning based on spatial relationships in images. The environments in ING-VP include six representative games: Sokoban, Maze, Sudoku, 8-queens, Tower of Hanoi, and 15-puzzle. In the evaluation of agents, three metrics are considered: accuracy, completion degree, and action efficiency.
DSGBench~\cite{tang2025dsgbench} provides a rigorous evaluation platform and includes six complex strategic games like StarCraft II, Civilization and so on to evaluate the agents on decision making. In all of the environments, the original observations—such as screenshots—are transformed into text prompts for the agents. For each game, there exist numerous human-designed metrics that are employed to assess the performance of agents. For instance, in StarCraft II, the following metrics are adopted: RPM (Resource Per Minute), EER (Efficiency of Resource Utilization), SUR (Supply Usage Rate), TCR (Technology Completion Rate), APM (Actions Per Minute), EPM (Ecffective Actions Per Minute), WR (Win Rate) and GA (Grounding Accuracy).

Another branches of benchmarks try to compare the agents and their environments usually support multiple agents.
Gamebench~\cite{costarelli2024gamebench} evaluates strategic reasoning abilities of AI agents, where nine multi-player board games are served as the testbeds, including Air, Land, Sea (ALS), Arctic Scavengers (ARC) and so on. The Bradley-Terry system~\cite{BradleyTerry1952} is employed to calculate the scores of the agents in order to make a comparison of their relative abilities.
In a similar vein, the $\gamma$-Bench~\cite{huang2025competing} and Gtbench~\cite{duan2024gtbench} assess the performance of agents in a manner analogous to that of competing models, albeit within the context of game-theoretic environments.
GameArena~\cite{hu2024gamearena} is distinguished from prior multi-agent game environments in that it incorporates human participation and provides human feedback during the evaluation process. For instance, in the AI Akinator game of GameArena, the agents determine what object the user is thinking of. During the game, the user responds to the agents' questions with a "yes" or "no" response.
\section{\raisebox{-0.7ex}{\includegraphics[height=3ex]{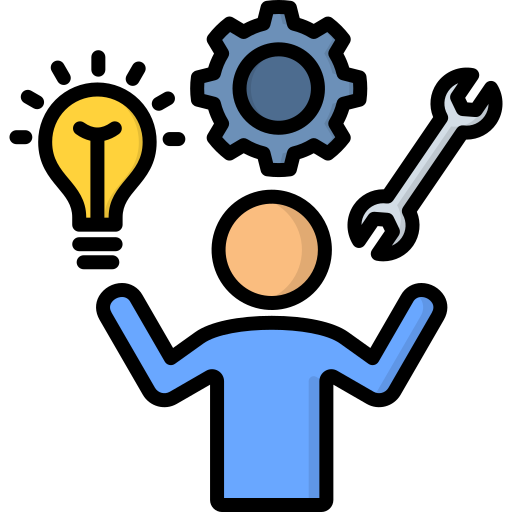}}~\textbf{Evaluation on Agent Capabilities}}
\label{sec:capabilities}

Evaluating AI agents necessitates a nuanced understanding of their advanced internal capabilities, which are fundamental to their performance across various tasks. This section categorizes existing evaluation methods based on these intrinsic abilities.

\subsection{\raisebox{-0.7ex}{\includegraphics[height=3ex]{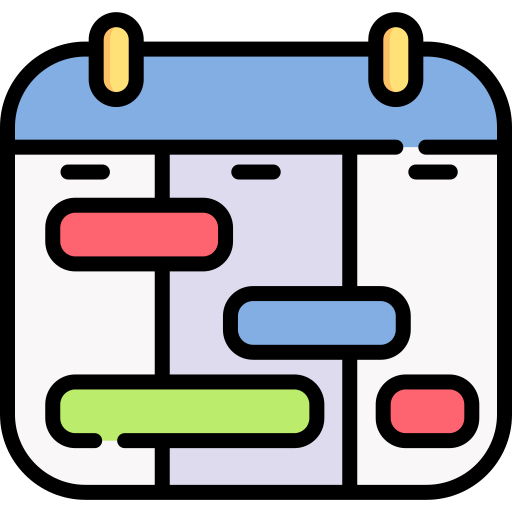}}~\textbf{Planning}}
\label{sec:planning}

As the complexity and diversity of environments confronted by AI agents increase, single‐step agents designed for isolated question answering fail to tackle tasks requiring multi‐step reasoning, such as solving advanced mathematical problems or navigating across multiple documents. To address these limitations, researchers have introduced methods to stimulate multi‐step reasoning in AI agents, including chain‐of‐thought prompting (CoT)~\citep{wei2022chain} and tree‐of‐thoughts (ToT)~\citep{yao2023tree} strategies.

The simplest approach to assess multi‐step reasoning is through static datasets spanning diverse domains. In mathematical reasoning, AQUA‐RAT~\citep{ling2017program} evaluates algebraic planning with generated rationales, GSM8K~\cite{cobbe2021training} and MATH~\cite{hendrycks2021measuring} cover elementary and secondary school problem solving, Game of 24~\citep{zhang2023cumulative} measures arithmetic CoT planning, and SVAMP~\citep{patel2021nlp} introduces perturbations such as operand swaps and paraphrases to increase difficulty on Math Word Problems. Document‐navigation benchmarks like HotpotQA~\cite{yang2018hotpotqa} and StrategyQA~\cite{geva2021did} require multi‐hop navigations across multiple passages. ScienceQA~\cite{lu2022learn} extends the standard multiple-choice science questions with multimodal information(e.g., charts and images), whereas ARC~\citep{clark2018think} consists of text-based multiple-choice science questions. Logic‐inference workflows are evaluated by FOLIO~\cite{han-etal-2024-folio} and P‐FOLIO~\citep{han2024p}, which treat logical inference chains as planning workflows.

Beyond implicit reasoning, explicit planning evaluation employs the Planning Domain Definition Language (PDDL) from classical AI planning. PlanBench leverages canonical domains such as Blocksworld and Logistics to test logical, causal, and spatial planning across eight subtasks \citep{valmeekam2023planbench}. \citet{valmeekam2023planning} provide a critical investigation showing agent planning success is limited but improves when used as heuristic guidance for external planners. AutoPlanBench~\cite{stein2023autoplanbench} automates the conversion of PDDL domains into natural language prompts, enabling large‐scale evaluation of agent planning methods and demonstrating that automatically generated prompts match or exceed manually crafted ones. ~\citet{bohnet2024exploring} introduces a benchmark suite covering both classical and natural‐language planning tasks, investigating many‐shot in‐context learning and fine‐tuning strategies to enhance planning performance. ACPBench~\cite{kokel2025acpbench} further decomposes planning into seven atomic reasoning tasks, such as action applicability, progression, reachability, and landmark detection, across thirteen domains to probe constraint-based planning abilities.

While traditional benchmarks focus on abstract or virtual tasks, workflow‐based benchmarks evaluate an agent's ability to decompose real‐world tasks into executable subtasks. NATURALPLAN~\cite{zheng2024natural} introduces realistic planning tasks like trip planning, meeting scheduling, and calendar management by providing tool outputs as context, eliminating the need for external execution environments. FlowBench~\cite{xiao2024flowbench} revisits workflow‐guided planning by formalizing multiple formats of workflow knowledge and covering 51 scenarios across six domains, revealing that current agents struggle with workflow hallucinations and require significant improvement. WorFBench~\citep{qiao2024benchmarking} offers a unified benchmark for workflow generation with complex graph structures and a systematic evaluation protocol utilizing subsequence and subgraph matching, highlighting distinct gaps between sequence and graph planning capabilities.

Finally, to incorporate environment feedback and iterative instruction, benchmarks such as MINT~\citep{wang2024mintevaluatingllmsmultiturn} and REALM‐Bench~\cite{geng2025realm} assess agents in interactive, online settings. MINT evaluates multi‐turn interaction capabilities by combining tool use and natural language feedback, showing that feedback can substantially improve performance but also revealing that instruction fine‐tuning may hurt multi‐turn reasoning. REALM‐Bench provides an online simulator with eleven real‐world planning scenarios, such as supply chain and disaster response, that can be scaled in parallelism, dependency complexity, and disturbance frequency, enabling rigorous testing of both single‐agent and multi‐agent planning under dynamic conditions.

\subsection{\raisebox{-0.7ex}{\includegraphics[height=3ex]{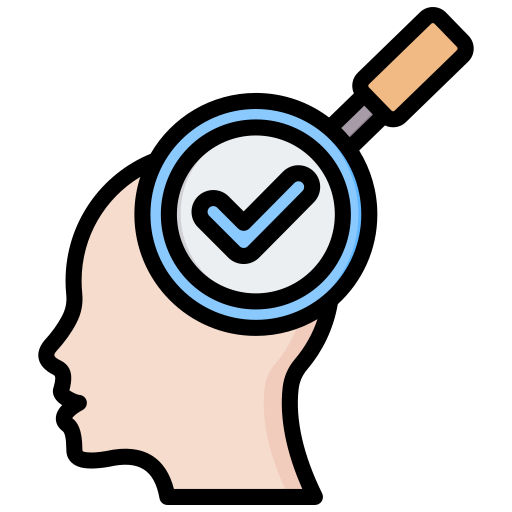}}~\textbf{Self-reflection}}
\label{sec:self-reflection}
% Self-reflection enables agents to assess and refine their actions and decisions, fostering continuous improvement. Evaluations focus on an agent's ability to introspect and adjust based on feedback. Key components include:

% \subsubsection{Error Detection}

% Identifying inaccuracies in responses or actions.

% \subsubsection{Learning from Feedback}: 

% Incorporating external feedback to enhance future performance.

As LLM chatbots evolve into more autonomous AI agents, the ability to self-reflect and improve through interaction—rather than relying solely on human supervision—has become increasingly critical. A growing body of research explores whether agents can engage in self-reflection through interactive feedback and subsequently refine their reasoning to reduce errors in multi-step tasks. This process demands not only an understanding of feedback but also the dynamic updating of internal beliefs, allowing agents to adaptively revise actions or reasoning paths over extended interaction trajectories.

Early evaluations of agent self-reflection are often simplistic and limited. These studies typically employ static environments with repetitive access and measure improvement only by checking whether the final answer is corrected after multiple attempts~\cite{Renze_2024,liu2025selfreflectionmakeslargelanguage}. However, such evaluations tend to be coarse and lack the capacity to generalize to new data or real-world feedback scenarios.

Furthermore, nascent benchmarks primarily focused on internal self-reflection, providing minimal environmental feedback and relying heavily on prompts such as 'Review your previous answer and find problems with your answer'~\cite{huang2024largelanguagemodelsselfcorrect,Renze_2024}. This internal-only mode of reflection limits the performance ceiling of agents, as relying solely on intrinsic reasoning cannot fully replace the adaptive learning signal provided by interactive rewards from the environment. Therefore, there is a pressing need for evaluation protocols that assess an agent’s ability to perform self-reflection grounded in external feedback signals.

To address this gap, LLF-Bench~\cite{cheng2023llfbenchbenchmarkinteractivelearning} was proposed as a standardized benchmark for interactive self-reflection. LLF-Bench constructs domain-specific language feedback across multiple tasks, enabling a systematic evaluation of AI agents' ability to learn from external linguistic corrections. The benchmark simulates diverse feedback scenarios and tests whether AI agents can adapt and improve based on natural language guidance. Similarly, LLM-Evolve~\cite{llm-evolve} targets the self-reflection ability of AI agents on standard benchmarks such as MMLU~\cite{mmlu}. In this framework, agents are exposed to dynamic feedback and allowed to retrieve few-shot examples for reflection and correction. The benchmark measures whether the integration of feedback leads to meaningful performance improvements across tasks.

Reflection-Bench~\cite{li2024reflectionbenchprobingaiintelligence} introduces novel evaluation metrics from a cognitive perspective to assess the quality, coherence, and logical consistency of reflective outputs. It explicitly examines whether a model can articulate the cause of its errors, propose plausible fixes, and ultimately achieve better decision-making as a result of reflection. Finally, \citet{pan2025benchmarkstalkreevaluatingcode} focuses on the domain of code generation, where self-reflection is evaluated in code environments based on human-provided feedback. Instead of only assessing whether a model generates correct code on the first try, the benchmark emphasizes the ability to revise and improve through iterative reflection, simulating real-world debugging and correction scenarios on three existing benchmarks APPS~\cite{APPS}, LiveCodeBench~\cite{jain2024livecodebenchholisticcontaminationfree}, and ClassEval~\cite{du2023classeval}.

\subsection{\raisebox{-0.7ex}{\includegraphics[height=3ex]{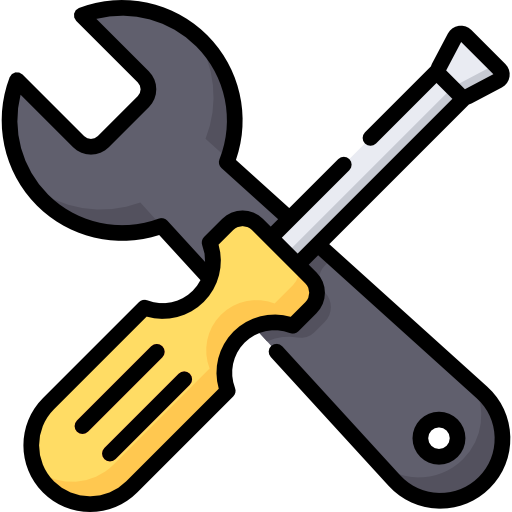}}~\textbf{Interaction}}
\label{sec:interaction}

To systematically evaluate the interaction capabilities of AI agents, we propose a three-level taxonomy of agent-environment interaction, each representing a progressively more complex and autonomous behavior. 

\subsubsection{\textbf{Interaction with Static Systems}}

First, we consider interaction with static systems, where the agent interacts with predefined tools or user interfaces (e.g., GUI APIs) via \textit{tool use} or \textit{function calling}. These interactions typically rely on a fixed set of functionalities. The focus of evaluation here is whether the agent can correctly interpret user intent, identify the appropriate tool, and execute the call correctly. This ability constitutes the foundation of agent competence.

Numerous benchmarks have been developed to assess the performance of tool agents across various tasks. These benchmarks evaluate aspects such as tool selection, parameter filling, output formatting, and the ability to handle complex, multi-step processes. A comprehensive overview of these benchmarks is presented in Table~\ref{tab:tooltable1} and Table~\ref{tab:tooltable2}. 

\textbf{General-Purpose Tool Agent Benchmarks}
Several benchmarks have emerged to evaluate agents' capabilities in handling a wide range of tool interactions. These tools involves a wide domain  as weather reports, news retrieval and translation. Benchmarks like ToolAlpaca~\cite{toolalpaca}, ToolBench~\cite{toolbench}, BFCL~\cite{BFCL}, Seal-Tools~\cite{sealtools} exemplify this phase. To be specific, early ones ToolAlpaca~\cite{toolalpaca} focuses on generalized tool learning by providing a diverse set of API interactions, and ToolBench~\cite{toolbench} is designed to evaluate agents' ability to manipulate real-world tools. BFCL~\cite{BFCL} maintains the quality and consistency of live documentation by updating  BFCLv2 and BFCLv3 to assess function calling capabilities. Seal-Tools~\cite{sealtools} employs self-instructed learning to generate tool usage scenarios. API-Bank~\cite{apibank} focuses on evaluating agents' ability to search and retrieve specific APIs. NexusRaven~\cite{nexusraven} assesses agents' performance in nested and parallel function calling scenarios. API-Blend~\cite{apiblend} emphasizes accurate slot filling and sequencing in complex scenarios.  APIGen~\cite{apigen} focuses on multi-turn function calling scenarios. StableToolBench~\cite{stabletoolbench} aims to provide stable and realistic tool evaluation by simulating real-world API interactions.

\textbf{Specialized Tool Agent Benchmarks}
In addition to general-purpose benchmarks, several specialized benchmarks have been developed to assess agents' capabilities of tool calling in specific domains. APIBench~\cite{apibench} challenges agents on evaluating the performance of agents in utilizing Python APIs for machine learning tasks. ToolSandBox~\cite{toolsandbox} evaluates agents' ability to interact with mobile app tools. Its primary difficulty is managing stateful interactions and implicit dependencies between tools.  RestBench~\cite{restbench} assesses agents' performance in interacting with movie and music databases. ComplexFuncBench~\cite{complexfuncbench} focuses on handling complex function calls, such as booking and reservation APIs while managing user constraints and preferences. NESTFUL~\cite{nestful} evaluates agents' ability to handle nested sequences of API calls, where outputs from one call serve as inputs to another. A core challenge is managing dependencies and ensuring correct sequencing. 

\subsubsection{\textbf{Interaction with Human}}

Second, we extend to interaction with humans, where the agent must engage in multi-turn interactions to accomplish a single goal, and humans exhibit changing states or behaviors. Agents have necessity to ask clarifying questions or request further instructions from a human user, due to the inherently dynamic and underspecified nature of human intent. 

Early benchmarks focused on constructing multi-turn human-agent dialogues manually to evaluate human-agent interactions like ABCD~\cite{abcd} and MultiWoZ~\cite{budzianowski2020multiwozlargescalemultidomain}. The Action-Based Conversations Dataset (ABCD) \cite{abcd} constructs dialogues requiring unique sequences of actions, constrained by company policies, to achieve task success. This dataset emphasizes the importance of aligning agent actions with organizational guidelines in realistic customer service scenarios. Similarly, the MultiWOZ dataset~\cite{budzianowski2020multiwozlargescalemultidomain} offers a large-scale collection of human-human written conversations spanning multiple domains. It serves as a valuable resource for developing and evaluating task-oriented dialogue systems across diverse contexts.

With the advent of AI agents, researchers began exploring the simulation of human users by agents instead of manual construction to enhance the realism of evaluation scenarios. The $\tau$-bench \cite{yao2024tau} benchmark emulates dynamic conversations between a user agent and a language agent equipped with domain-specific API tools and policy guidelines. This setup allows for the assessment of an agent's ability to interact with simulated users.

In recent years, benchmarks have begun to combine automated and manual processes to generate diver see and realistic dialogues. Specifically, ALMITA~\cite{ALMITA} proposed a new dialogue dataset and framework specifically aimed at evaluating tool-augmented dialogue AI agents in customer support scenarios. IntellAgent~\cite{yao2024intellagent}  employs a graph-based representation to model the relationships, likelihoods, and complexities of various policies. This approach enables the simulation of multi-policy scenarios, capturing the nuanced interplay between agent capabilities and policy constraints.

We have systematically compiled these works into a comparative Table~\ref{tab:human_agent_interaction}. From this table, it becomes evident that the primary task in human-agent interaction benchmarks is often tool calling. Consequently, the ability to effectively utilize tools serves as the cornerstone of agent interaction capabilities. Subsequent interactions with humans are typically aimed at obtaining feedback or further instructions to enhance the completion of tool-based tasks.

% \jc{Are these two paragraphs necessary?}
% However, we contend that current research in human-agent interaction evaluation remains insufficient. Firstly, the prevailing benchmarks predominantly focus on task accuracy, assessing only the final completion of tasks without directly evaluating the quality of human-agent interactions. For instance, they do not measure whether the questions posed by agents are pertinent, whether a query is necessary, or if the language used is polite. We expect the future incorporation of intermediate milestones to provide a more intuitive assessment of the questions agents pose during interactions.

% Secondly, existing benchmarks are predominantly offline, with agents interacting with fixed datasets or user simulators, lacking real-time interactions with actual humans. This absence of genuine human interaction diminishes the ecological validity of the evaluations. Lastly, current benchmarks primarily utilize text-based modalities. In contrast, real-world human feedback encompasses not only textual input but also non-verbal cues such as facial expressions (images) and tone of voice (audio). The absence of these multimodal feedback channels in evaluations overlooks critical aspects of human-agent interactions.

\subsubsection{\textbf{Interactions with Other Agents}}
\label{sec:multi-agent-interactions}

Once agents acquire the capability to independently solve problems, a third mode of interaction emerges, which is agent-agent interaction. The purpose of this interaction is twofold: to collaboratively accomplish tasks that a single agent cannot achieve alone, or to engage in competitive scenarios to determine a winner. To systematically evaluate these interactions, we categorize existing benchmarks into two paradigms: \textbf{cooperative} and \textbf{competitive}. These paradigms serve as the foundation for assessing agent-agent interactions, as shown in Table~\ref{tab:agent-agent-interactions}.

According to cooperative agent evaluations, Collab-Overcooked~\cite{Wang2025CollabOvercooked} extends the Overcooked-AI game to evaluate AI agents' collaborative capabilities by incorporating process-oriented evaluation metrics such as Trajectory Efficiency Score (TES) and Incremental Trajectory Efficiency Score (ITES). SmartEvals~\cite{Alahi2023SmartEval} encompasses six distinct games, including Rock-Paper-Scissors, Tower of Hanoi, and Minecraft, each augmented with language descriptors for visual observation. The benchmark uniquely challenges nine critical capabilities of intelligent AI agents, such as reasoning with object dependencies, planning ahead, spatial reasoning, learning from history, and understanding randomness. MindAgent~\cite{Yu2024MindAgent} leverages existing gaming frameworks to require not only understanding of a multi-agent system but also collaboration with human players via un-finetuned instructions. GOVSIM~\cite{Piatti2024GOVSIM} presents scenarios such as fishery, pasture, and pollution, focusing on agents engaged in cooperative behavior and effective governance to achieve sustainable outcomes 

For competitive agents evaluation, AutoArena~\cite{Zhao2024AutoArena} is an automated evaluation framework that assesses AI agents through peer battles and committee discussions. It utilizes an agent examiner to generate questions, followed by multi-round peer battles between agent candidates, and concludes with a committee of agent judges collaboratively deciding the winner. This approach reduces bias and enhances evaluation fairness 

% \mh{two recently? except competitive or cooperative xxx}
Recently, researchers have established multi-agent benchmarks that simultaneously consider cooperation and competition. BattleAgentBench~\cite{Wang2024BattleAgentBench} evaluates AI agents across seven sub-stages with three varying difficulty levels. It assesses single-agent navigation, paired-agent task execution, and multi-agent collaboration and competition capabilities. MultiAgentBench~\cite{Gangrade2025MultiAgentBench} captures coordination dynamics and competitive interactions, providing tailored metrics such as Key Performance Indicators (KPIs), structured planning and communication scores, and competition scores to systematically assess agent performance. Sotopia~\cite{Wang2023SOTOPIA} simulates complex social interactions between artificial agents and humans. It evaluates agents' social intelligence through role-play interactions under various scenarios, assessing dimensions like believability, relationship, knowledge, secrecy, social rules, financial/material benefits, and goal completion.

\subsection{\raisebox{-0.7ex}{\includegraphics[height=3ex]{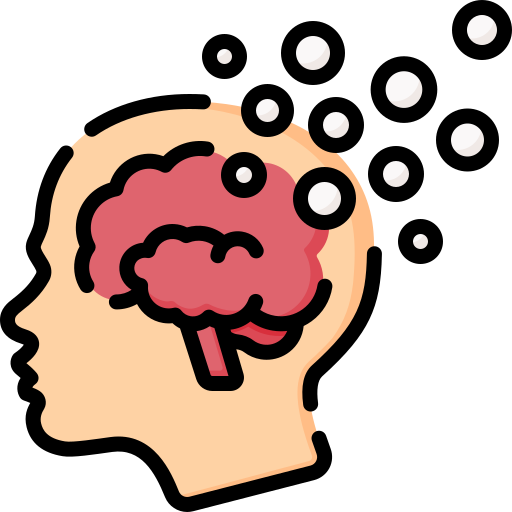}}~\textbf{Memory}}
\label{sec:memory}

Memory encompasses an agent's ability to store, retrieve, and apply information over time. A robust memory system allows agents to maintain coherence in long interactions, retrieves relevant knowledge when needed, and adapts behavior based on past experiences. As memory becomes an increasingly vital component for complex reasoning and decision-making tasks, several benchmarks have been developed to systematically assess different aspects of agent memory. We summarize these benchmarks in Table~\ref{tab:memory_benchmarks}.

Some benchmarks focus on evaluating specific memory capabilities, such as question answering (QA) or summarization. NarrativeQA~\cite{kovcisky2018narrativeqa} tests an agent’s ability to answer questions based on long books or screenplays, emphasizing deep narrative understanding. QMSum~\cite{zhong2021qmsum} evaluates memory through the task of generating multi-domain meeting summaries from lengthy transcripts. QuALITY~\cite{pang2021quality} targets reading comprehension, challenging agents to answer multiple-choice questions that require understanding and recalling extended documents. DialSim~\cite{kim2024dialsim} focuses on dialogue-based QA for TV series, requiring agents to maintain and reason over dialogue history and spatiotemporal memory.

Other benchmarks adopt a more holistic approach, examining multiple memory skills or diverse content types. LoCoMo~\cite{maharana2024evaluating} assesses agents through question answering, event summarization, and multimodal dialogue generation tasks. LTM-Benchmark~\cite{castillo2024beyond} evaluates conversational agents’ long-term memory and information integration within a single dynamic multitask dialogue. LongMemEval~\cite{wu2024longmemeval} tests agents across information extraction, multi-session reasoning, temporal reasoning, knowledge updates, and abstention. Episodic Memory Benchmark~\cite{huet2025episodic} assesses episodic memory capabilities, focusing on spatiotemporal-contextual event recall, cross-event relational reasoning, and entity state tracking. PerLTQA~\cite{du2024perltqa} evaluates agents' ability to integrate and apply semantic and episodic memories in personalized long-term memory question answering. Finally, StreamBench~\cite{wu2024streambench} focuses on online, streaming learning, assessing an agent’s capacity to incrementally update its memory and refine performance based on feedback streams.

\subsection{\raisebox{-0.7ex}{\includegraphics[height=3ex]{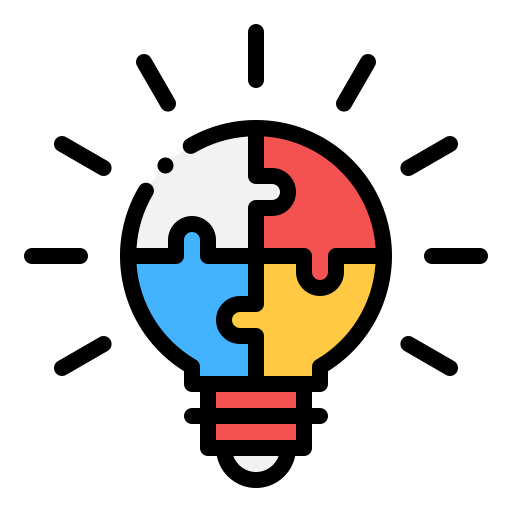}}~\textbf{General}}
\label{sec:general agent}
Beyond the aforementioned agent benchmarks that assess performance in specific capabilities, there exist several benchmarks that explicitly evaluate multiple abilities, including planning and memory. These benchmarks enable a holistic assessment of an agent's generalization capabilities, transcending the limitations of particular environments or individual skills.

AgentBench~\cite{agentbench} pioneered the realm of general capabilities benchmarks. It is the first multi-dimensional evaluation framework for AI agents, encompassing eight distinct environments, including operating systems, databases, and games. The benchmark evaluates 27 commercial LLMs, revealing performance variations across complex scenarios. Similarly, GAIA~\cite{mialon2023gaiabenchmarkgeneralai} and MMAU~\cite{mmau2024} also offer offline evaluation benchmarks. GAIA focuses on real-world assistant tasks. It consists of 466 human-designed and annotated questions that are text-based and may include files like images or spreadsheets. These questions are intended to reflect real-world challenges. GAIA evaluates AI systems against real-world tasks through these questions, testing fundamental abilities like reasoning, multi-modality handling, web browsing, and tool-use proficiency. MMAU constructs five major domains encompassing tool usage, question answering, mathematics, and machine learning coding. It assesses AI agents from multiple perspectives, including reasoning, planning, problem-solving, and self-correction.

As research progresses, the limitations of offline evaluations, such as timeliness and lack of realism, have prompted the development of online evaluation benchmarks. Notably, Galileo's Agent Leaderboard~\cite{galileo_agent_leaderboard2023} introduces a novel evaluation framework that assesses language models' proficiency in real-world agentic tasks by measuring Tool Selection Quality (TSQ), offering a comprehensive ranking across multiple domains such as mathematics, entertainment, education, and retail. HAL (Holistic Agent Leaderboard)~\cite{hal} provides a standardized, cost-aware platform for evaluating AI agents, integrating a unified evaluation harness that supports various benchmarks and facilitates reproducible assessments, thereby enabling transparent comparisons across different agent implementations. Additionally, AgentArena~\cite{agentarena2025}, akin to ChatbotArena~\cite{chiang2024chatbot}, allows human evaluators to anonymously score two agents, facilitating authentic preference assessments. This approach not only provides evaluations but also generates human preference data, fostering a positive feedback loop for further agent training.

Furthermore, several innovative general agent benchmarks have emerged. Agent-as-a-Judge~\cite{agent_as_a_judge2024} departs from traditional human evaluations by employing agents to assess other agents, thereby streamlining the evaluation process and mitigating the cognitive limitations inherent in human assessments. We consider this a promising trend, enhancing scalability in agent evaluations and addressing the challenges associated with human evaluators, which will be discussed in Section~\ref{sec:discussion}. Lastly, Capabench~\cite{whos_the_mvp2025} utilizes Shapley values to analyze the contributions of different modules within an agent across various tasks and environments, offering insights into the modular performance and optimization of AI agents.

\section{\textbf{Discussion: The Evolution of Agent Evaluation}}
\label{sec:discussion}

\begin{figure*}[t]
  \centering
  \vspace{-10pt}
  \includegraphics[width=1.0\textwidth]{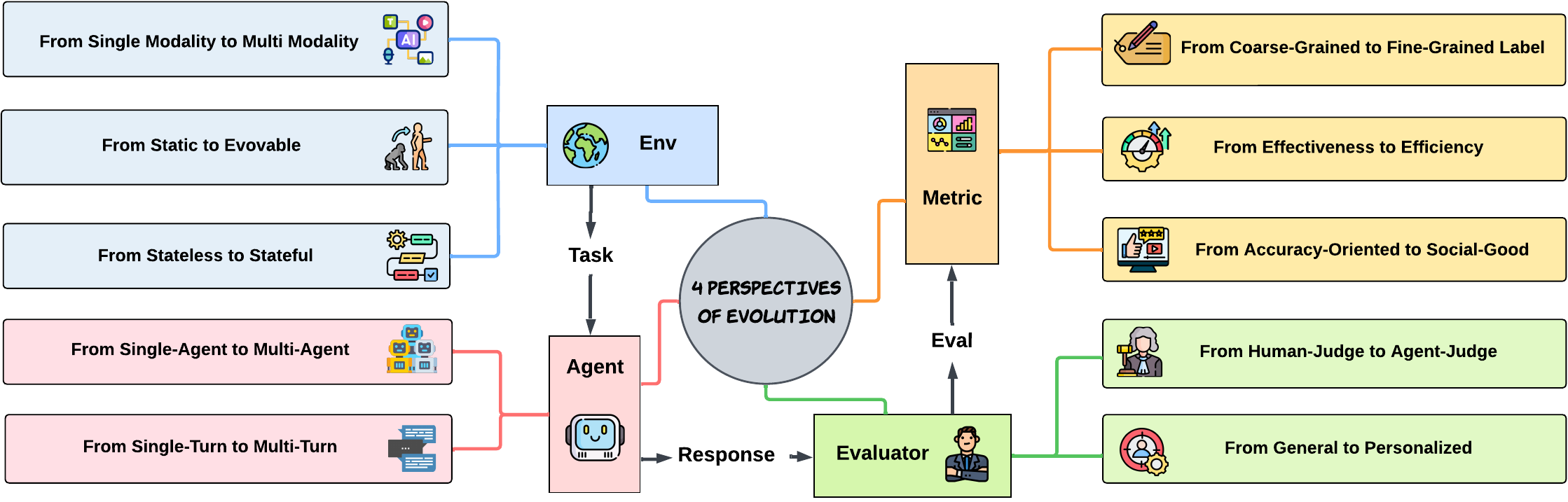}
  \vspace{-10pt}
  \caption{Four perspectives of evolution prospect about AI agent evaluation.}
  \label{fig:discussion}
  \vspace{-10pt}
\end{figure*}

After presenting a variety of agent evaluation methods and benchmarks, a natural question arises: \textit{Given the rapid advancement of AI agents, how can we systematically evaluate LLM-based AI agents from an evolutionary perspective?} This question will be answered in the final section of this discussion. In this section, we will discuss the evolution of agent evaluation from four perspectives, providing both a summary of past work and guidance for future research.

The evolutionary framework is shown in Figure~\ref{fig:discussion}, where we can observe the evaluation process and its four key modules. First, a task is extracted from the environment for the agent to solve. Then, the agent produces a response, which is evaluated by an evaluator through processes such as calculation and judgment. Finally, various evaluation metrics are generated. We will explore each of these four modules in detail to discuss the future directions of Agent Evaluation.

\subsection{\textbf{Environment Perspective}}
\label{sec:env perspective}

\textbf{From Single Modality to Multi Modality}.
Early AI agents were primarily text-based, and consequently, most early benchmarks focused on evaluating agents using text-based inputs. This was reasonable at the time. As shown in Tables~\ref{tab:tooltable1} and~\ref{tab:human_agent_interaction}, most benchmarks involving tools, APIs, and human-agent interactions relied on text. However, as the complexity of environments has increased, we are now witnessing the introduction of additional modalities, such as images in web environments or chat audio in mobile environments. Studies have demonstrated that agents perform better when they have access to visual information~\cite{visualwebarena}. Thus, incorporating multimodal evaluation is crucial and beneficial. We are already seeing multimodal benchmarks related to images and videos, such as VisualAgentBench~\cite{liu2024visualagentbenchlargemultimodalmodels} and VideoWebArena~\cite{visualwebarena}, and we believe this will be the future trend. Interestingly, in gaming environment benchmarks (e.g., SmartPlay~\cite{wu2023smartplay}, \(\tau\)-bench~\cite{yao2024tau}), while game images are present, benchmarks have chosen to represent the positions of all characters using coordinates, converting images into text-based inputs for agents. We believe that future evaluations will involve more direct image input rather than converting it into text.

\textbf{From Static to Evolving.} 
Earlier benchmarks are relatively simple, where researchers collect data and publish it in offline systems for evaluation. Such benchmarks are static. However, in the real world, the environment evolves rapidly, particularly on the Internet, where new information emerges daily. If benchmarks are not regularly updated, they risk becoming obsolete. Under this philosophy, creating evolving, scalable benchmarks is essential. One approach is to manually update benchmarks, as seen with BFCL~\cite{BFCL}, BFCLv2, BFCLv3, and the swe-bench~\cite{zan2408swe} series. By continually injecting new knowledge into benchmarks, they can stay relevant. Another approach involves connecting to real-time online evaluations, such as webvoyage~\cite{he2024webvoyager} and WorkArena~\cite{workarena}. These benchmarks are highly scalable because they are directly connected to the ever-evolving Internet, maintaining their relevance and expansion capabilities.

\textbf{From Stateless to Stateful.}
% \rrt{The title is also confusing—it sounds like "from stateless to stateful environments", but seems to be "from stateless to stateful evaluation" from the paragraph?}
As agent evaluation evolves, there is a shift from stateless evaluations to stateful ones. Early benchmarks typically provide isolated, single-instance evaluations. However, agents in the real world interact with complex, evolving states. Thus, incorporating stateful evaluation, where the agent's past interactions and context influence its performance, is increasingly important. This approach allows for more accurate evaluations of an agent's ability to handle dynamic and persistent tasks over time, rather than just isolated scenarios.

Earlier benchmarks, especially in web or mobile environments, are generally stateless, focusing only on the agent's final state. Most benchmarks initialize the agent to a starting state, required the agent to complete a task, and then assessed whether it reached the final state (e.g., a booking confirmation page or a retrieved piece of information). However, the environments in these benchmarks were limited in their ability to evaluate the intermediate states of the agent’s process. As the field progresses, there is a shift toward considering the entire sequence of an agent’s actions, not just the final outcome. Some benchmarks, such as ToolSandBox~\cite{toolsandbox} and OmniAct~\cite{OmniACT}, now assess intermediate states by introducing milestones or calculating a sequence score instead of just the final result. This evolution towards stateful evaluation is increasingly relevant in real-world tasks where agents must consider multiple intermediate states, such as handling interruptions in complex workflows. For example, in a multi-task scenario like "book a flight in 20 minutes, organize a document, and search for information simultaneously," agents need to manage both task-specific states and more complex intermediate states. Recent benchmarks, like Worfbench~\cite{qiao2024benchmarking}, have already begun addressing such multi-task inputs, signaling a broader trend from stateless to stateful evaluation.

\subsection{\textbf{Agent Perspective}}
\label{sec:agent perspective}

\textbf{From Single-Agent to Multi-Agent.}
Most current benchmarks focus on single-agent tasks, where the agent operates independently to solve a problem. However, as the field matures, the focus is shifting toward multi-agent systems, where agents collaborate or compete to achieve their respective goals. Multi-agent environments introduce new challenges in coordination, communication, and negotiation, all of which require novel benchmarking approaches. We believe multi-agent benchmarks hold significant potential. Currently, they face two key challenges.

Primarily, current benchmarks predominantly feature game-based or virtual tasks, lacking scenarios reflective of real-world collaborative activities, such as internet-based teamwork. Benchmarks like MultiAgentBench~\cite{Gangrade2025MultiAgentBench} and BattleAgentBench~\cite{Wang2024BattleAgentBench}, while valuable for evaluating coordination and competition, do not adequately represent practical collaborative contexts.

Furthermore, most existing benchmarks focus on homogeneous multi-agent systems, neglecting the essential role of heterogeneous agents with specialized skills—such as reasoning (Deepseek~\cite{guo2025deepseek}), retrieval (GPT~\cite{openai2023gpt4}), and coding (Claude~\cite{anthropic2025claude})—and hierarchical structures commonly seen in real-world settings (e.g., manager-executer dynamics). Although frameworks like DeepClaude~\cite{deepclaude2025} and AgentVerse~\cite{agentverse} have begun exploring these specialized and hierarchical collaborations, standardized benchmarks for evaluating heterogeneous agent teams remain underdeveloped.

\textbf{From Single-Turn to Multi-Turn.}
In the early stages of AI agent evaluation, benchmarks are often designed for single-turn interactions, where the agent receives a prompt and responds immediately. Tasks like single-turn retrieval or end-to-end task execution require minimal interaction. Benchmarks like WebArena~\cite{webarena} are designed for such scenarios. As agent capabilities improve, agents now engage in multi-turn dialogues to refine user needs or receive further instructions. Benchmarks like $\tau$-Bench~\cite{yao2024tau} or Weblinx~\cite{lu2024weblinx} now accommodate these extended interactions, and we believe the future of agent evaluation will increasingly involve multi-turn conversations. This trend will naturally lead to the emergence of benchmarks that evaluate the ability to manage long contextual dialogue sequences.

\subsection{\textbf{Evaluator Perspective}}
\label{sec:evaluator perspective}

\textbf{From Human-Judge to Agent-Judge.}
Initially, humans are seen as the best evaluators for agents, given that agents were designed to assist humans with tasks. Early evaluations, such as WebArena~\cite{webarena} or AgentArena~\cite{agentarena2025}, rely on human judges to assess agent performance. However, over time, it becomes apparent that using agent-based evaluators (agent-judges) could offer significant advantages. First, agent-judges can scale effectively without requiring additional human resources. Second, as AI agents' capabilities evolve and surpass human performance in some domains, human evaluation may no longer align with the agents' full potential. As such, using agents to evaluate other agents is increasingly seen as a necessary evolution in the field.

\textbf{From General to Personalized.}
Currently, most agents are designed to be general-purpose, aimed at completing tasks based on fixed instructions. Benchmarks reflect this approach, with tasks such as retrieving information or booking tickets. However, once agents are deployed in real-world commercial applications, they must consider factors like user preferences, profiles, and personal history. Personalized agent benchmarks, such as PetoolBench~\cite{petoolbench}, have emerged to evaluate how well agents adapt to individual users. These benchmarks will play a crucial role as personalized agents become more common in commercial settings.

\subsection{\textbf{Metric Perspective}}
\label{sec:metric perspective}

\textbf{From Coarse-Grained to Fine-Grained Label.}
Most benchmarks provide coarse-grained evaluations, offering broad feedback based on the final task outcome. However, a more granular evaluation system is necessary to capture the nuances of an agent's performance. Fine-grained labels, such as textual comments on an agent’s decision-making process (e.g., “this agent’s planning incurs high time costs” or “this agent’s decisions are fast but risky”), could provide more valuable insights to developers. Although such fine-grained feedback may not be suitable for directly comparing agents by score or ranking which one is better, it can provide developers with more valuable insights, enabling the creation of improved agents.

\textbf{From Effectiveness to Efficiency.}
While effectiveness metrics—such as task completion or accuracy—have traditionally been the focus of agent evaluation, efficiency is becoming an increasingly important consideration. How quickly and resource-efficiently an agent performs a task is essential. Benchmarks like Worfbench~\cite{qiao2024benchmarking} and GOVSIM~\cite{Piatti2024GOVSIM} already assess efficiency alongside effectiveness, and future benchmarks will need to balance both aspects to ensure that agents complete tasks efficiently, without overconsuming resources or time.

 \textbf{From Accuracy-Oriented to Social-Good.}
As AI agents are integrated into society, evaluating their alignment with societal values such as fairness, safety, and ethical behavior is becoming a critical focus. Current benchmarks have begun to incorporate safety metrics, such as those found in ST-WebAgentBench~\cite{st-webagentbench}. However, additional metrics, such as trustworthiness, robustness, and the ethical implications of agents’ actions (e.g., making erroneous payments or generating harmful content), are still lacking and should be developed further in future benchmarks. We hope that agents will contribute to society in a socially beneficial way, rather than causing any harmful issues.

\subsection{\textbf{Benchmark Selection Methodology}}
\label{sec:benchmark_selection_methodology}
Having detailed the evolution of agent evaluation from four distinct perspectives, we now return to address the core question posed at the beginning of our discussion section: \textit{Given the rapid advancement of AI agents, how can we systematically evaluate LLM-based AI agents from an evolutionary perspective?} To tackle this question, we propose a two-stage benchmark selection methodology containing present-focused Selection and future-oriented selection, and illustrate it vividly through a practical use case.

\textbf{Step 1: Present-focused Selection.}
This step addresses the immediate evaluation needs that arise when developers have newly created an agent requiring immediate assessment. For example, consider the developer Z, who has developed an initial agent capable of booking flights and hotels. In this scenario, the developer can first consult Figure~\ref{fig:agent-evaluation-typology} to identify relevant benchmark categories by considering the agent’s external environment and internal capabilities. For this particular use case, the environment is web-based, and the primary capability involves interaction, specifically tool calling. Subsequently, the developer Z may proceed to the corresponding category sections to obtain an overview of each benchmark. Crucially, the developer could examine the associated category table provided in the Appendix~\ref{sec:app}, which enumerates attributes we deem essential for informed benchmark selection such as modality, task domain, data source, and evaluation metrics. Based on these attributes, the developer Z can identify suitable benchmarks. In this use case scenario, benchmarks such as WebVoyager~\cite{he2024webvoyager} (for web environments) and ComplexFuncBench~\cite{complexfuncbench} (for interaction capabilities) would be appropriate choices.

\textbf{Step 2: Future-oriented Selection.}
This second step considers the prospective evolution and ongoing improvement of agents, addressing the potential shifts in evaluation methodologies required as the agent matures. Developers can leverage Figure~\ref{fig:discussion}, which provides a high-level prospective view on the evolution of agent evaluation. This figure assists developers in two critical areas: (1) identifying directions for further optimization of their agents, and (2) anticipating the future dimensions along which their agents might be assessed. Continuing our previous use case, the external environment might evolve dynamically, necessitating continuous attention to benchmarks adaptable to environmental changes exemplified by BFCL. Similarly, evaluation metrics may shift from purely task-oriented accuracy towards socially beneficial considerations like safety and robustness, reminding the developer Z of monitoring benchmarks like ST-WebAgentBench~\cite{st-webagentbench}. Additionally, as the developer Z's product moves towards commercialization, incorporating user history, greater attention to personalized benchmarks such as PeToolBench~\cite{petoolbench} will become increasingly important.

Through these two methodological steps, coupled with the illustrative use case provided, developers can systematically select suitable benchmarks for their agents. Furthermore, this approach equips developers with a high-level perspective on future evaluation trends, aiding in the continuous enhancement and comprehensive assessment of agent performance.

\section{\textbf{Conclusion}}
% This paper presents a comprehensive framework for evaluating AI agents, distinguishing them from traditional LLM chatbots through five key dimensions: Perception, Instructor, Capability, Environment, and Feedback. We systematically categorize existing evaluation benchmarks based on external environments (e.g., coding, web, gaming) and internal capabilities (e.g., planning, memory, interaction), providing a structured approach to benchmark selection.
% We also highlight the evolutionary trends in AI agent evaluation across environment, agent, evaluator, and metric dimensions, offering guidance for future research. As AI agents continue to evolve, our work provides a foundation for advancing evaluation methods and benchmarks, helping researchers navigate the growing complexity of these systems and select benchmarks for their own agents.

In this paper, we have systematically explored the evolutionary transition from traditional LLM chatbots to advanced AI agents, highlighting critical evolution across five aspects: complex environment, multi-source instructor, dynamic feedback, multi-modal perception, and advanced capability. By adopting an evolutionary perspective, we have addressed existing ambiguities in evaluation practices and offered a structured categorization of benchmarks aligned with external environments and internal capabilities. Our comprehensive attribute-based tables serve as practical resources for researchers aiming to select suitable evaluation frameworks. Furthermore, we have discussed evolving trends and provided forward-looking insights on evaluation methodologies from four perspectives of environment, agent, evaluator, and metric. This work not only clarifies the evaluation landscape for AI agents but also sets the stage for future research directions, ultimately contributing to the ongoing advancement and refinement of AI agent systems.

% if have a single appendix:
%\appendix[Proof of the Zonklar Equations]
% or
%\appendix  % for no appendix heading
% do not use \section anymore after \appendix, only \section*
% is possibly needed

% use appendices with more than one appendix
% then use \section to start each appendix
% you must declare a \section before using any
% \subsection or using \label (\appendices by itself
% starts a section numbered zero.)
%

% \appendices
% \section{Proof of the First Zonklar Equation}
% Appendix one text goes here.

% you can choose not to have a title for an appendix
% if you want by leaving the argument blank
% \section{}
% Appendix two text goes here.

% use section* for acknowledgment
% \section*{Acknowledgment}

% The authors would like to thank...

% Can use something like this to put references on a page
% by themselves when using endfloat and the captionsoff option.
\ifCLASSOPTIONcaptionsoff
  \newpage
\fi

% trigger a \newpage just before the given reference
% number - used to balance the columns on the last page
% adjust value as needed - may need to be readjusted if
% the document is modified later
%\IEEEtriggeratref{8}
% The "triggered" command can be changed if desired:
%\IEEEtriggercmd{\enlargethispage{-5in}}

% references section

% can use a bibliography generated by BibTeX as a .bbl file
% BibTeX documentation can be easily obtained at:
% http://mirror.ctan.org/biblio/bibtex/contrib/doc/
% The IEEEtran BibTeX style support page is at:
% http://www.michaelshell.org/tex/ieeetran/bibtex/
%\bibliographystyle{IEEEtran}
% argument is your BibTeX string definitions and bibliography database(s)
%\bibliography{IEEEabrv,../bib/paper}
%
% <OR> manually copy in the resultant .bbl file
% set second argument of \begin to the number of references
% (used to reserve space for the reference number labels box)
% \begin{thebibliography}{1}

% \bibitem{IEEEhowto:kopka}
% H.~Kopka and P.~W. Daly, \emph{A Guide to \LaTeX}, 3rd~ed.\hskip 1em plus
%   0.5em minus 0.4em\relax Harlow, England: Addison-Wesley, 1999.

% \end{thebibliography}
\bibliographystyle{IEEEtranN}
\bibliography{IEEEabrv,./bibtex/bib/ref}

% biography section
% 
% If you have an EPS/PDF photo (graphicx package needed) extra braces are
% needed around the contents of the optional argument to biography to prevent
% the LaTeX parser from getting confused when it sees the complicated
% \includegraphics command within an optional argument. (You could create
% your own custom macro containing the \includegraphics command to make things
% simpler here.)
%\begin{IEEEbiography}[{\includegraphics[width=1in,height=1.25in,clip,keepaspectratio]{mshell}}]{Michael Shell}
% or if you just want to reserve a space for a photo:

% \begin{IEEEbiography}{Michael Shell}
% Biography text here.
% \end{IEEEbiography}

% % if you will not have a photo at all:
% \begin{IEEEbiographynophoto}{John Doe}
% Biography text here.
% \end{IEEEbiographynophoto}

% % insert where needed to balance the two columns on the last page with
% % biographies
% %\newpage

% \begin{IEEEbiographynophoto}{Jane Doe}
% Biography text here.
% \end{IEEEbiographynophoto}

% You can push biographies down or up by placing
% a \vfill before or after them. The appropriate
% use of \vfill depends on what kind of text is
% on the last page and whether or not the columns
% are being equalized.

%\vfill

% Can be used to pull up biographies so that the bottom of the last one
% is flush with the other column.
%\enlargethispage{-5in}

\newpage
\appendices

\section{Tables of Taxonomy}
\label{sec:app}

To assist researchers in efficiently and accurately selecting appropriate evaluation benchmarks for various types of AI agents, we have systematically organized the majority of the categories identified in our taxonomy into tables. Specifically, for each benchmark category, we extract and summarize the most valuable attributes that characterize its intrinsic properties and suitability for different evaluation scenarios. These attributes were carefully selected to guide researchers in clearly understanding the strengths and limitations of each benchmark, thus providing structured guidance for evaluating AI agents and benchmark selection.

Below, we present the summarized tables for each benchmark category:

\subsection{\textbf{Coding Environments}}

\begin{table*}[h]
\centering   
% \vspace{-10pt}
\caption{Code Benchmarks and Their Taxonomy}

% \vspace{-8pt}
% \hspace{-30pt}
\resizebox{1.0\textwidth}{!}{
\renewcommand\arraystretch{1.2}
% \begin{tabular}{lp{2cm}p{2cm}p{2cm}p{4cm}p{3cm}p{3cm}p{3cm}}
\begin{tabular}{p{3cm}p{2.7cm}p{5cm}p{2.5cm}p{4cm}}
\hline
Code Environments/Datasets &  \textbf{Static/Interactive}  & \textbf{Tasks} & \textbf{Evaluation Metric} & \textbf{Evaluation Dimensions} \\  
\hline 

HumanEval\cite{chen2021evaluating} & Static & Code generation from docstring & Pass@k & Code correctness\\ \hline 
Classeval\cite{du2023classeval} & Static & Class-level code generation & Pass@k, Method/Field Dependencies & Code correctness, Independency to the contexts\\ \hline 
\citet{yuan2023evaluating} & Static & Defect detection, Clone detection, Assert generation, Code summarization & Accuracy, F1 score, EM(Exact Match) & Correctness of detection, assert and summarization\\ \hline 
CodeCriticBench\cite{zhang2025codecriticbench} & Static & Code generation, Code QA & Accuracy, MSE, Pass@1 & Code correctness, Code evaluation accuracy\\ \hline 
CRUXEval\cite{gu2024cruxeval} & Static & Code input and output prediction & Pass@k & Code prediction correctness\\ \hline 
BigCodeBench\cite{zhuo2024bigcodebench} & Static & Code generation with tools & Pass@k & Code correctness\\ \hline 
ARCADE\cite{yin2022natural} & Static & Code generation in interactive data science notebooks & Pass@k & Code correctness\\ \hline 
SWE-bench\cite{jimenez2023swe} & Interactive & Code generation for resolving the issues in the repository & Resolved Rate & Correctness of patch files\\ \hline 
Swe-bench M\cite{yang2024swe} & Interactive & Code generation for resolving the issues in the repository with images & Resolved Rate, Average cost & Correctness of patch files\\ \hline 
SWE-bench-java-verified\cite{zan2408swe} & Interactive & Code generation for resolving the issues in the repository & Resolved Rate & Correctness of patch files\\ \hline 
Swe arena\cite{swe-arena2024} & Interactive & Code generation for resolving the issues in the repository & Resolved Rate & Correctness of patch files\\ \hline 
Swe-bench+\cite{aleithan2024swe} & Interactive & Code generation for resolving the issues in the repository & Resolved Rate, Average Cost & Correctness of patch files\\ \hline 
RepoBench\cite{liu2023repobench} & Interactive & Code retrieval and code completion on code repositories & Accuracy, Exact Match, Edit Similarity & Accuracy of code retrieval, Accuracy of code completion\\ \hline 
SWT-Bench\cite{mundler2024swt} & Interactive & Code generation for resolving the issues in the repository & Success Rate, Line Coverage,  & Success rate, Change coverage, Patch well-Formedness\\ \hline 
DevEval\cite{li2024prompting}  & Interactive & Software design, environment setup, implementation and testing & LLM-as-a-judge, Success/Pass Rate & General principles and faithfulness for software design, Correctness for environment setup, Correctness of implmentation, Correctness and code coverage of testing\\ \hline 
ML-Bench\cite{tang2023ml} & Interactive & Machine learning tasks like package installing, model code generation & Pass@k & Completion of machine learning tasks\\ \hline 
Pybench\cite{zhang2024pybenchevaluatingllmagent} & Interactive & Python code generation for Chart Analysis, Text Analysis, Image \& Audio Editing, Complex Math and Software \& Website Development & Pass Rate, Average Turns, LLM-as-a-judge & Completion of python related tasks \\

\hline
\end{tabular}

\label{tab:code Env}
\vspace{-5pt}
}
\end{table*}

We categorize benchmarks of coding environments in Table~\ref{tab:code Env}. The columns in Table~\ref{tab:code Env} have the following meanings:
\begin{itemize}[leftmargin=10pt]
    \item \textbf{Code Environments/ Datasets}: Refers to the specific benchmarks or datasets used to evaluate AI agents or language models on coding-related tasks.
    \item \textbf{Static/ Interactive}: Specifically classifies each benchmark as either static (isolated tasks without ongoing interaction) or interactive (tasks requiring continuous engagement with an environment or repository).
    \item \textbf{Tasks}: Describes the primary coding tasks evaluated by each benchmark, such as code generation, defect detection, retrieval, summarization, or software engineering tasks.
    \item \textbf{Evaluation Metric}: Lists the metrics used to quantify model performance, including common metrics like Pass\@k, Accuracy, Resolved Rate, Edit Similarity, and specialized indicators like Method/Field Dependencies.
    \item \textbf{Evaluation Dimensions}: Clarifies the specific aspects or dimensions being assessed by each benchmark, for example, correctness of code, independency from context, accuracy in defect detection, or the overall completion quality of tasks.
    
\end{itemize}

\subsection{\textbf{Web Environments}}

\begin{table*}[h]
\centering   
% \vspace{-10pt}
\caption{Web Benchmarks and Their Taxonomy}

% \vspace{-8pt}
% \hspace{-30pt}
\label{tab:Web Env}
\resizebox{1.0\textwidth}{!}{
\renewcommand\arraystretch{1.2}
% \begin{tabular}{lp{2cm}p{2cm}p{2cm}p{4cm}p{3cm}p{3cm}p{3cm}}
\begin{tabular}{p{3cm}p{1.6cm}p{1.5cm}p{3cm}p{3cm}p{2cm}p{4.5cm}}
\hline
Web Environment &  \textbf{Realism}  & \textbf{Offline/Online} & \textbf{Metrics} & \textbf{Evaluator} & \textbf{Input Modalities} & \textbf{Main Challenges} \\  
\hline 

MiniWoB~\cite{miniwob}    & Synthetic  & Offline &Task Success Rate & Human labels + Automated evaluation & Text + Images & A reproducible offline RL benchmark for open‑domain web tasks. \\ \hline 
FormWoB~\cite{miniwob}   & Semi‑real  & Offline & Average Reward & Human labels + Automated evaluation & Text + Images & Same as above. \\ \hline 
QAWoB~\cite{miniwob}      & Semi‑real & Offline & Test Trajectory Likelihood & Human labels + Automated evaluation & Text + Images & Same as above. \\ \hline 
MiniWoB++~\cite{miniwob++}  & Synthetic & Offline &Task Success Rate & Human labels + Automated evaluation & Text + Images & Added longer sequences, random layouts, and soft‑text reasoning tasks. \\ \hline 
WebShop~\cite{webshop}    & Semi‑real  & Offline & Independent scoring on attributes/options/price/type & Human labels + Automated evaluation & Text + Images & Requires agents to complete full purchase flow  \\ \hline 
mind2web~\cite{mind2web}   & Real  (snapshot replay) & Offline & Element Accuracy, Operation F1 & Human labels + Automated evaluation & Text + Images & Very large number of web pages. \\ \hline 
webvoyage~\cite{he2024webvoyager}  & Real (live connection)   & Online & Human vs GPT‑4V Agreement & Human +LLM joint evaluation & Text + Images & Online real environment judged directly by humans. \\ \hline 

WEBLINX~\cite{lu2024weblinx}           & Real (snapshot replay)              & Offline & IoU, URL‑F1                                 & Human labels + Automated evaluation & Text + Images     & Introduces “conversational web navigation”\\ \hline 
webarena~\cite{webarena}          & Semi‑real    & Offline & Intermediate page states                    & Human labels + Automated evaluation & Text + Images     & Self‑hosted multi‑site environment that closely mirrors real websites. \\ \hline 
visualwebarena~\cite{visualwebarena}    & Semi‑real       & Offline & Task Success Rate                           & Human labels + LLM‑based similarity & Text + Images     & Visual‑dependent tasks: “see then do” scenarios. \\ \hline 
WorkArena~\cite{workarena}         & Real (live connection)                 & Online  &Task Success Rate                                 & Fully automated by real web         & Text + Images     & Runs on official ServiceNow developer instance; fully realistic. \\ \hline 
WorkArena++~\cite{workarena++}       & Real (live connection)                & Online  & Task Success Rate                            & Fully automated by real web      & Text + Images     & Evaluates “compositional planning and reasoning.” \\ \hline 
MMInA~\cite{MMInA}             & Real (live connection)                 & Online  & Hop Success Rate                             & Human labels + LLM‑based similarity & Text + Images     & Multi‑hop (cross‑site chain operations). \\ \hline 
AssistantBench~\cite{assistantbench}    & Real (live connection)      & Online  & Task Success Rate            & Human labels + LLM‑based similarity & Text + Images     & Time‑consuming, multi‑step workflows. \\ \hline 
WebCanvas ~\cite{webcanvas}        & Real  & Online  & Task Success Rate                            & Human labels + Automated evaluation & Text + Images     & Continuously updated with robust intermediate checks. \\ \hline 
ST‑WebAgentBench~\cite{st-webagentbench}  & Semi‑real        & Online  & Completion‑under‑Policies, Risk Ratio          & Human labels + Automated evaluation & Text + Images     & Policy compliance focus under risk constraints. \\ \hline 
VideoWebArena~\cite{videowebarena}     & Semi‑real   & Online  & Video QA Accuracy                            & Human labels + Automated evaluation & Video + Text + Images & Long‑video + web interaction (Skill vs Factual tasks). \\ \hline 
TUR[K]INGBENCH~\cite{xu2025turkingbenchchallengebenchmarkweb}    & Semi‑real      & Offline & ROUGE‑L, IoU                                & Human labels + Automated evaluation & Text + Images     & Natural HTML pages originally designed for crowdsourcing. \\ \hline 
BEARCUBS~\cite{bearcubs}          & Real     & Online  & error rate  & Human labels + Automated evaluation & Video + Text + Images & Fine‑grained answer labeling with live multimedia items. \\ \hline 
THEAGENTCOMPANY~\cite{xu2024theagentcompanybenchmarkingllmagents}   & Semi‑real    & Offline & Full‑Completion Score                         & Human + Automated + LLM similarity    & Text + Images     & Performing real company tasks. \\ \hline 
WABER~\cite{kara2025waber}          & Semi‑real      & Online  & Reliability, Efficiency                      & Human labels + Automated evaluation & Text + Images     & Injected failures + efficiency benchmarking. \\ \hline 
VisualAgentBench~\cite{liu2024visualagentbenchlargemultimodalmodels}  & Semi‑real      & Offline & Task Success Rate                            & Human labels + Automated evaluation & Text + Images     & Multi‑domain, multi‑scenario LMM visual agent benchmark. \\ \hline 
Online‑Mind2Web~\cite{online-mind2web}   & Real (live connection)   & Online  & Task Success Rate                            & Human labels + LLM‑based similarity & Text + Images     & Online adaptation of Mind2Web. \\ \hline 
REAL~\cite{REAL}              & Semi‑real     & Offline & Rubric‑Guided Retrieval Score                 & Human labels + LLM/Rubric similarity    & Text + Images     & Tests both robustness and reliability simultaneously. \\ \hline 
OmniAct~\cite{OmniACT}           & Real       & Offline & Sequence Score, Action Score                 & Human labels + Automated evaluation & Text + Images     & Desktop + web automation with DetACT module. \\ \hline 
ChatShop~\cite{chatshop}          & Semi‑real    & Offline & RL Learning Curves (ChatShopBin only)        & Human labels + Automated evaluation & Text + Images     & Ambiguous initial info requiring strategic questioning. \\

\hline
\end{tabular}
\vspace{-5pt}
}
\end{table*}

We categorize benchmarks of web environments in Table~\ref{tab:Web Env}. The columns in Table~\ref{tab:Web Env} have the following meanings:
\begin{itemize}[leftmargin=10pt]
    \item \textbf{Web Environment}: Identifies the specific benchmarks or datasets designed for evaluating AI agents in web-based tasks or scenarios.
    \item \textbf{Realism}: Describes how closely the evaluation environment replicates real-world conditions, ranging from synthetic (highly controlled or artificial), semi-real (partially realistic), to fully real environments.
    \item \textbf{Offline/Online}: Specifies whether the benchmark tasks are conducted offline (pre-recorded, simulated data or snapshot replay) or online (live, interactive connections with dynamic web content).
    \item \textbf{Metrics}: Lists the evaluation metrics used to quantify agent performance, including task success rate, trajectory likelihood, element accuracy, operation success, and other task-specific indicators.
    \item \textbf{Evaluator}: Indicates how evaluation results are obtained, whether through human annotators, automated systems, or combined human and automated evaluation methods, including similarity to LLM-generated judgments.
    \item \textbf{Input Modalities}: Clarifies the types of input data provided to the agents, typically involving textual descriptions and images, or in some cases video and additional media.
    \item \textbf{Main Challenges}: Summarizes the key difficulties or unique features presented by each benchmark, providing insight into the specific aspects that make these web-based evaluations challenging and relevant for assessing AI agents.

\end{itemize}

These columns collectively help researchers quickly understand the scope, complexity, realism, and evaluation approach of each web benchmark, guiding the selection of appropriate benchmarks for assessing web-based AI agents.

\subsection{\textbf{OS Environments}}
\begin{table*}[h]
\centering   
% \vspace{-10pt}
\caption{OS Benchmarks and Their Taxonomy}
% The symbol $\ast$ indicates a statistically significant improvement of RetrievalPRM over the best baseline with $p$-value < 0.01.}
% \vspace{-8pt}
% \hspace{-30pt}
\label{tab:os_benchmarks}
\resizebox{1\textwidth}{!}{
\begin{tabular}{p{3cm}p{3cm}p{3cm}p{2cm}p{2cm}p{3cm}}
\hline
Benchmark &  \textbf{Tasks}  & \textbf{Observation} & \textbf{Action Space} & \textbf{Platform}& \textbf{Evaluation Method}\\  
\hline 

OSWorld\cite{xie2024osworld} & Office tasks, OS tasks, Daily tasks, Workflow tasks and Professional tasks & Screenshots, a11y-tree & Keyboard, Mouse & Ubuntu & Predefined post-processing scripts to calculate success rate \\ \hline 
WindowsAgentArena\cite{bonatti2024windows} & Office, Web browsing, Windows system, Coding, Media \& Video, Windows utilities & Task instruction, Clipboard content, Metadata of the current session, Representation of the current screen & Keyboard, Mouse & Windows & Predefined post-processing scripts to calculate success rate\\ \hline 
AgentStudio\cite{zheng2024agentstudio} & Office tasks, OS tasks, Daily tasks and so on & Screen recording, Screenshots, Code execution outputs & Keyboard, Mouse, APIs/Tools & Linux, (Windows, macOS) & User-defined functional auto-evaluators based on reward and language feedback\\ \hline 
OmniACT\cite{kapoor2024omniact} & Shopping tasks, Entertainment tasks, Service tasks, Government tasks, Travel tasks, Health tasks & Screenshots & Actions in PyAutoGUI library & Linux, Windows, macOS & Action sequences matching, Inaccurate behavior penalty\\ \hline 
OfficeBench\cite{wang2024officebench} & Office tasks, like Word, Excel, PDF, Email and so on & Text description of current state and previous outputs of actions & APIs & Simulated desktop & Exact matching and fuzzy matching on the final outputs, Scripts to check the final state\\ \hline 
PC-Eval\cite{liu2025pc} & Commom tasks on Chrome, Word, Excel and so on & Screenshots, Operation histories & Keyboard, Mouse, APIs & Windows & Human evaluation on success rate and subtask success rate \\

\hline
\end{tabular}
% \vspace{-5pt}
}
\end{table*}

We categorize benchmarks of OS environments in Table~\ref{tab:os_benchmarks}. The columns in Table~\ref{tab:os_benchmarks} have the following meanings:

\begin{itemize}[leftmargin=10pt]
    
    \item \textbf{Tasks}: Lists the main types of tasks the AI agents are expected to perform within these OS benchmarks, such as office-related activities, system operations, web browsing, coding tasks, and daily workflow tasks.

    \item \textbf{Observation Type}: Details the format and type of information provided to the agent for decision-making or interactions, such as screenshots, accessibility trees (a11y-tree), screen recordings, clipboard content, metadata, or textual descriptions.

    \item \textbf{Action Space}: Defines the set of actions or interfaces available to the agent, including input devices like keyboards and mice, APIs, or specialized GUI automation libraries (e.g., PyAutoGUI).
    \item \textbf{Platform}: Indicates the specific operating systems or platforms supported by each benchmark, such as Ubuntu, Windows, Linux, macOS, or simulated desktop environments.
    \item \textbf{Evaluation Method}: Describes the approach used to measure agent performance, including predefined scripts for automatic success rate calculation, action sequence matching, human evaluation, exact or fuzzy output matching, and penalties for inaccurate behaviors.
\end{itemize}

Collectively, these columns clearly outline the characteristics, operational details, and evaluation strategies of OS benchmarks, guiding researchers in selecting suitable platforms for assessing AI agents within realistic operating system contexts.

\subsection{\textbf{Mobile Environments}}
\begin{table*}[ht]

\caption{Mobile Benchmarks and Their Taxonomy}
\centering
\label{tab:mobile}
\resizebox{1.0\textwidth}{!}{
\begin{tabular}{{p{3cm}p{2cm}p{3.5cm}p{4cm}p{2cm}p{1.2cm}p{4cm}}}
\hline
\textbf{Benchmark} & \textbf{Static/Executable} & \textbf{Tasks} & \textbf{Observation Type} & \textbf{Action Space Type} & \textbf{Platform} & \textbf{Evaluation Method} \\ \hline

PixelHelp\cite{li2020mapping} & Static & Instructions following & Text view hierarchy & Screen operations & Android & Action sequences match \\ \hline 
UGIF\cite{venkatesh2022ugif} & Static & Instructions following & Text view hierarchy & Screen operations + Macro Functions & Android & Action sequences match \\ \hline 
MoTIF\cite{burns2021mobile} & Static & Instruction following and instruction feasible classification & Screenshots + Text view hierarchy & Screen operations & Android & Calculate F1-score on the binary classification problem \\ \hline 
AITW\cite{rawles2023androidinthewild} & Static & Instructions following & Screenshots + Screen representation & Screen operations & Android & Action sequences match + Human validation \\ \hline 
ANDROIDCONTROL\cite{li2024effects} & Static & Instructions following & Screenshots + Text view hierarchy & Screen operations + Macro functions & Android & Action sequences match \\ \hline 
AMEX\cite{chai2024amex} & Static & Instructions following & Screenshots + Screen representation (description, element grounding) & Screen operations & Android & Action sequences match \\ \hline 
B-MOCA\cite{lee2024benchmarking} & Executable & Instructions following & Screenshots + Text view hierarchy & Screen operations & Android & Self-developed success detector with ADB and Appium \\ \hline 
Androidworld\cite{rawles2024androidworld} & Executable & Instructions following & Screenshots + Text view hierarchy & Screen operations + APIs & Android & Inferring task success with ADB \\ \hline 
Mobile-bench\cite{deng2024mobile} & Executable & Instructions following & HTML converted from XML & Screen operations + APIs & Android & Checkpoints + GPT4 evaluate task completion status \\ \hline 
A3\cite{chai2025a3} & Executable & Instructions following & Screenshots + Text view hierarchy & Screen operations & Android & Task-specific evaluation function (element matching, action matching) + LLM Evaluation System \\ \hline 
Mobile-Env\cite{zhang2023mobile} & Executable & Instructions following & Screenshots + Text view hierarchy + Set-of-Marks & Screen operations & Android & Evaluation from screen text,  VH, system log and RHU \\ \hline 
AndroidArena\cite{xing2024understanding} & Executable & Cross-APP instructions following & Screenshots + Text view hierarchy & Screen operations + APP-level actions + System-level actions & Android & Action sequences match + GPT4 judgement on success \\ \hline 
MobileSafetyBench\cite{lee2024mobilesafetybench} & Executable & Safety-related instructions following (e.g. Finance, Device/Data Management) & Screenshots + Text view hierarchy &  Screen operations & Android & Rule-based evaluators on goal achievement and harm prevention \\ \hline 
MobileAgentBench\cite{wang2024mobileagentbench} & Executable & Instructions following & Screenshots + Text view hierarchy & Screen operations & Android & Final UI state matching with self-developed app \\ \hline 
SPA-BENCH\cite{chen2024spa} & Executable & Single-APP and Cross-APP instructions following & Screenshots + Text view hierarchy & Screen operations & Android & Key components matching  + MLLM evaluation \\ \hline 
Llamatouch\cite{zhang2024llamatouch} & Executable & Instructions following & Screenshots + Text view hierarchy & Screen operations & Android & Evaluating on whether the agent  traverses all manually annotated application/system state \\ \hline 
Androidlab\cite{xu2024androidlab} & Executable & Instructions following & Screenshots + Text view hierarchy & Screen operations & Android & Evaluate on the completion of sub-goals \\ \hline 
AutoEval\cite{sun2025autoeval} & Executable & Instructions following & Screenshots + Text view hierarchy & Screen operations & Android & LLM-based automatic judge system, including Capturer, Reasoner and Checker \\
\hline
\end{tabular}

}
\end{table*}

We categorize benchmarks of mobile environments in Table~\ref{tab:mobile}. The columns in Table~\ref{tab:mobile} have the following meanings:
\begin{itemize}[leftmargin=10pt]
    \item \textbf{Static/Executable}: Indicates whether the benchmark consists of static scenarios (tasks based on pre-defined and static interfaces or screens) or executable scenarios (dynamic, real-time interactions with mobile applications).
    \item \textbf{Tasks}: Specifies the core task evaluated by each benchmark, predominantly focusing on instruction-following capabilities within mobile interfaces.
    \item \textbf{Observation Type}: Details the types of visual or textual information provided to the agent, such as screenshots, text view hierarchies, screen representations, HTML/XML content, and additional marks or metadata.
    \item \textbf{Action Space Type}: Defines the range of actions available to the agent, typically involving screen operations (e.g., tapping, swiping) and, in some benchmarks, macro functions or API calls.
    \item \textbf{Platform}: Denotes the mobile operating system for which the benchmark is intended, predominantly Android.
    \item \textbf{Evaluation Method}: Describes how performance is measured, using methods such as action sequence matching, LLM-assisted automatic judgments, rule-based evaluation, self-developed success detection, human validation, or UI state matching.

\end{itemize}

These columns collectively provide a clear overview of mobile benchmarks, highlighting how tasks are structured, how agent observations are presented, and how agent actions and performance are evaluated, thus guiding researchers in selecting appropriate mobile environment evaluation for AI agents.

\subsection{\textbf{Scientific Environments}}

\begin{table*}[h]
\centering   
% \vspace{-10pt}
\caption{Scientific Benchmarks and Their Taxonomy}
\label{tab:scientific}

% \vspace{-8pt}
% \hspace{-30pt}
\resizebox{1.0\textwidth}{!}{
\renewcommand\arraystretch{1.2}
% \begin{tabular}{lp{2cm}p{2cm}p{2cm}p{4cm}p{3cm}p{3cm}p{3cm}}
\begin{tabular}{lllll}
\hline
Scientific Benchmarks &  \textbf{\makecell[l]{Static/ \\Interactive}}  & \textbf{Environment} & \textbf{Evaluation Metric} & \textbf{Tasks} \\  
\hline 

ARC~\cite{clark2018think} & Static & None & Accuracy & Scientific question answering\\ \hline 
SCIENCEQA~\cite{lu2022learn} & Static & None & Accuracy & Multimodal scientific question answering\\ \hline 
QASPER~\cite{dasigi2021dataset} & Static & None & Span-level F1 score & Research paper information seeking and answer anchoring\\ \hline 
QASA~\cite{lee2023qasa} & Static & None & Precision, recall, F1 score, ROUGE scores & Question answering on scientific articles\\ \hline 
SWIFT~\cite{chamoun2024automated} & Static & None & Specificity, actionability, reading
comprehension, overall helpfulness & Feedback generation for scientific writing\\ \hline 
AAAR-1.0~\cite{lou2024aaar} & Static & None & \makecell[l]{Equation inference: Accuracy;\\ Other tasks: human expert ratings} & \makecell[l]{Research assistance: equation inference, paper weakness,\\  experiment design, review critique}\\ \hline 
$MS^2$~\cite{deyoung2021ms2} & Static & None & $\Delta$EI & Multi-document summarization of medical studies\\ \hline 
LAB-Bench~\cite{laurent2024lab} & Static & None & Accuracy, precision, coverage & Multi-choice questions on biology research\\ \hline 
ResearchArena~\cite{kang2024researcharena} & Static & \makecell[l]{Academic and\\ survey papers} & \makecell[l]{Information retrieval: recall, precision; \\Information selection: nDCG, MRR;\\ Information Organization: Heading Entity Recall, Heading Soft Recall} & Academic surveying\\ \hline 
~\cite{si2024can} & Static & Research paper pool & Novelty, excitement, feasibility, expected effectiveness & Research idea generation\\ \hline 
PaperQA~\cite{lala2023paperqa} & Static & Scientific article pool & Retrieval AUC, retrieval probability & Scientific article question answering\\ \hline 
SCIENCEWORLD~\cite{wang2022scienceworld} & Interactive & \makecell[l]{Simulated text\\ environment} & \makecell[l]{Task completion rate, step success rate, \\average completion steps, error rate, learning curve} & Elementary-level scientific reasoning\\ \hline 
DISCOVERYWORLD~\cite{jansen2406discoveryworld} & Interactive & \makecell[l]{Simulated text\\ environment} & \makecell[l]{Task completion rate, task-relevant actions taken,\\ discovered explanatory knowledge} & Perform complete cycles of novel scientific discovery\\ \hline 
ScienceAgentBench~\cite{chen2024scienceagentbench} & Interactive & Workspace  & Valid execution rate, success rate, codeBERTScore, API Cost  & Data-driven scientific discovery\\ \hline 
MLGym~\cite{nathani2025mlgym}  & Interactive & Workspace & Area Under the Performance Profile (AUP) score & Realistic research workflows\\ \hline 
MLAgentBench~\cite{huang2310mlagentbench} & Interactive & Workspace & Success rate, consumed tokens and time & Machine learning experimentation tasks\\ \hline 
DSBench~\cite{jing2024dsbench} & Interactive & Workspace & Success rate, relative performance gap & Complex data science tasks \\ \hline 
DSEval~\cite{zhang2024benchmarking} & Interactive & Workspace & Accuracy, relative performance gap & \makecell[l]{Data analysis tasks and data modeling tasks \\throughout the entire data science lifecycle} \\ \hline 
DA-Code~\cite{huang2024code} & Interactive & Workspace & Accuracy & Data science code generation \\ \hline 
ML-Bench~\cite{tang2023ml} & Interactive & Workspace & Pass@k, success rate & Repository-level code understanding and execution tasks \\ \hline 
Spider2-V~\cite{cao2024spider2} & Interactive & Workspace & Success rate & Automate data science and engineering workflows \\

\hline
\end{tabular}
\vspace{-5pt}
}
\end{table*}

We categorize benchmarks of scientific environments in Table~\ref{tab:scientific}. The columns in Table~\ref{tab:scientific} have the following meanings:
\begin{itemize}[leftmargin=10pt]
    
    \item \textbf{Static/Interactive}: Indicates whether the benchmark involves static scenarios (predefined, non-interactive tasks) or interactive environments (requiring agents to perform tasks dynamically, often in simulated or real workspaces).
    \item \textbf{Environment}: Describes the context or type of environment in which the benchmark tasks are set, such as research papers, scientific article pools, survey data, or simulated workspaces.
    \item \textbf{Evaluation Metric}: Specifies the criteria and metrics used to assess agent performance, such as accuracy, F1 score, precision, recall, coverage, retrieval probability, area under the curve (AUC), success rate, error rate, and other task-specific indicators.
    \item \textbf{Tasks}: Details the main scientific or research-oriented tasks evaluated by each benchmark, including scientific question answering, research paper summarization, feedback generation, equation inference, data modeling, code generation, data analysis, and full-cycle scientific discovery.
\end{itemize}
Together, these columns provide a structured overview of scientific benchmarks, clarifying their operational format, application context, evaluation methodologies, and task focus, to guide researchers in choosing appropriate tools for assessing AI agents in scientific research scenarios.

\subsection{\textbf{Game Environments}}

\begin{table*}[h]
\centering   
% \vspace{-10pt}
\caption{Game Benchmarks and Their Taxonomy}

% \vspace{-8pt}
% \hspace{-30pt}
\label{tab:game Env}
\resizebox{1.0\textwidth}{!}{
\renewcommand\arraystretch{1.0}
% \begin{tabular}{lp{2cm}p{2cm}p{2cm}p{4cm}p{3cm}p{3cm}p{3cm}}
\begin{tabular}{p{2cm}p{4cm}p{1.5cm}p{6cm}}
\hline
Benchmark &  \textbf{Games}  & \textbf{Single-agent/Multi-agent} & \textbf{Needed Agent Capabilities} \\  
\hline 

BALROG\cite{paglieri2024balrog} & BabyAI, TextWorld, Crafter, Baba Is AI, MiniHack, NLE & Single-agent & Navigation, Exploration, Resource Management, Complex Credit Assignment, Deducing Env. Dynamics, Long-term Planning \\ \hline 
SmartPlay\cite{wu2023smartplay} & Bandits, Rock-Paper-Scissors, Hanoi, Messenger, Crafter, Minecraft & Single-agent & Long Text Understanding, Multi-hop Reasoning, Instruction/Rule Following, Planning, Generalization, Understanding the Odds, Learning from Interactions, Error/Mistake Handling, Spatial Reasoning\\ \hline 
LVLM-Playground \cite{wang2025large} & TicTacToe, Reversi, Sudoku, Minesweeper, Gomoku, Chess & Single-agent & Perception, Reasoning, Decision, Adversary\\ \hline 
VGRP-Bench\cite{ren2025vgrp} & Binairo, Star-Battle, Colored-Sudoku, Killer-Sudoku (20 visual reasoning puzzles) & Single-agent & Reasoning, Rule-following, Overall Perception, Cell-level Perception\\ \hline 
ING-VP\cite{zhang2024ing} & Sokoban, Maze, Sudoku, 8-queens, Tower of Hanoi, 15-puzzle & Single-agent & Perception, Reasoning, Planning, Text Understanding, Spatial Imagination\\ \hline  
DSGBench\cite{tang2025dsgbench} & Starcraft II, Civilization, Diplomacy, Stratego, Street Fighter III, Werewolf & Single-agent & Strategic Planning, Real-Time Decision Making, Adaptive Learning, Social Reasoning, Team Collaboration\\ \hline 
Gamebench\cite{costarelli2024gamebench} & Air, Land, Sea, Arctic Scavenger,  Are You the Traitor?, Codenames, Hive, Pit, Santorini, Two Rooms and a Boom, Sea Battle & Multi-agent & Reasoning capabilities on Abstract Strategy, Non-Deterministic, Hidden Information, Language Communication, Social Deduction and Cooperation\\ \hline 
$\gamma$-Bench\cite{huang2025competing} & Guess 2/3 of the Average, El Farol Bar, Divide the Dollar, Public Goods Game, Diner’s Dilemma, Sealed-Bid Auction, Battle Royale, Pirate Game & Multi-agent & Reasoning capabilities on game theoretic environments\\ \hline 
Gtbench\cite{duan2024gtbench} & Tic-Tac-Toe, Connect-4, Kuhn Poker, Breakthrough, Liar's Dice, Blind Auction, Negotiation, Nim, Pig, Iterated Prisoner's Dilemma & Multi-agent & Board Strategy, Bids, Collaboration, Bluff, Math\\ \hline 
GameArena\cite{hu2024gamearena} & Akinator, Taboo, Bluffing & Multi-agent & Deductive reasoning, Abductive reasoning, Inductive reasoning, Multi-hop reasoning\\

\hline
\end{tabular}
\vspace{-5pt}
}
\end{table*}

We categorize benchmarks of game environments in Table~\ref{tab:game Env}. The columns in Table~\ref{tab:game Env} have the following meanings:
\begin{itemize}
    \item \textbf{Games}: Specifies the types of games or specific game titles included within each benchmark, ranging from simple puzzles and board games to complex, strategic, or multiplayer games.
    \item \textbf{Single-agent/Multi-agent}: Indicates whether the benchmark focuses on evaluating single-agent scenarios (where an agent operates independently) or multi-agent scenarios (involving collaboration, competition, or interaction among multiple agents).
    \item \textbf{Needed Agent Capabilities}: Describes the key cognitive and behavioral abilities required from agents to succeed in the respective benchmark. These include skills such as reasoning, planning, decision-making, exploration, handling hidden information, cooperation, communication, and various forms of complex or multi-hop reasoning.
\end{itemize}

Together, these columns provide a comprehensive overview of game benchmarks, clarifying the context, interaction modality, and agent competencies needed, thereby assisting researchers in selecting appropriate gaming environments for evaluating diverse agent capabilities.

\subsection{\textbf{Planning Capability}}
\begin{table*}[h]
\centering
\caption{Planning Benchmarks and Their Taxonomy}
\label{tab:planning benchmarks}
\resizebox{\textwidth}{!}{
\renewcommand\arraystretch{1.2}
\begin{tabular}{lp{3cm}p{3cm}p{3cm}p{5cm}}
\hline
\textbf{Benchmark} & \textbf{Domain Category} & \textbf{Task Type} & \textbf{Format} & \textbf{Key Focus} \\
\hline
AQUA-RAT~\citep{ling2017program}                  
  & Mathematical Reasoning       & Algebraic word problem solving with natural-language rationales      & Static dataset             & Evaluating generated rationales in algebra problems \\
\hline
GSM8K~\citep{cobbe2021training}                    
  & Mathematical Reasoning       & Elementary arithmetic word problem solving (CoT)         & Static dataset     & Evaluating CoT performance on basic arithmetic problems \\
\hline
MATH~\citep{hendrycks2021measuring}                
  & Mathematical Reasoning       &  Competition-level math contest problem solving      & Static dataset             & Evaluating multi-step reasoning on challenging competition problems \\
\hline
Game of 24~\citep{zhang2023cumulative}            
  & Mathematical Reasoning       & Cumulative arithmetic game problem solving               & Static dataset             & Evaluating cumulative multi-step arithmetic reasoning to reach a target value \\
\hline
SVAMP~\citep{patel2021nlp}                        
  & Mathematical Reasoning       & Math word problem solving with simple variations         & Static dataset             & Evaluating brittleness of SOTA models under simple textual variations \\
\hline
HotpotQA~\citep{yang2018hotpotqa}                  
  & Document Navigation          & Multi-hop question answering                    & Static dataset             & Evaluating cross-document multi-hop reasoning \\
\hline
StrategyQA~\citep{geva2021did}                    
  & Document Navigation          & Implicit multi-hop inference          & Static dataset             & Evaluating implicit reasoning for strategy questions \\
\hline
ScienceQA~\citep{lu2022learn}                     
  & Science Knowledge            & Multimodal multiple-choice            & Static dataset             & Incorporating charts and images \\
\hline
ARC~\citep{clark2018think}                        
  & Science Knowledge            & Text-based multiple-choice            & Static dataset             & Purely textual science questions including both Easy and Challenge subsets\\
\hline
FOLIO~\citep{han-etal-2024-folio}                 
  & Logic Inference              & Natural-language reasoning with first-order logic annotations             & Static dataset             &  Evaluating NL reasoning and NL→FOL translation with formal proof verification \\
\hline
P-FOLIO~\citep{han2024p}                          
  & Logic Inference              & Step-by-step annotated natural-language proofs for FOL reasoning          & Static dataset             & Assessing and improving multi-step logical reasoning in LLMs through human-written proofs \\
\hline
PlanBench~\citep{valmeekam2023planbench}          
  & Classical Planning           & Planning in canonical IPC domains (Blocksworld \& Logistics)                 & PDDL (Static)              & Evaluating eight core planning capabilities (generation, optimality, verification, execution reasoning, robustness, reuse, replanning, generalization) \\
\hline
On Planning Abilities of LLMs~\citep{valmeekam2023planning} 
  & Classical Planning           & Autonomous plan generation and heuristic-guided planning         & PDDL + LLM prompts         & Evaluating LLM’s autonomous planning success and its use as heuristic guidance for external planners \\
\hline
AutoPlanBench~\citep{stein2023autoplanbench}      
  & Classical Planning           & Automated NL prompt generation from PDDL domains      & NL prompts(Static)               & Comparing auto-generated vs.\ manual and template-based NL prompts; enabling large-scale PDDL planning evaluation \\
\hline
Exploring Planning Capabilities~\citep{bohnet2024exploring} 
  & Planning Evaluation          & PDDL planning (BlocksWorld, Logistics, Mini-Grid); NL planning (Trip Planning, Calendar Scheduling)        & PDDL \& NL prompts (Static)               & Automated instance generation; many-shot in-context learning; chain-of-thought reasoning; fine-tuning; out-of-distribution generalization; failure mode analysis \\
\hline
ACPBench~\citep{kokel2025acpbench}                
  & Planning Evaluation          & Seven atomic planning reasoning tasks: Applicability, Progression, Reachability, Validation, Action Reachability, Justification, Landmark Detection                & PDDL-based NL prompts (Static)              & Benchmarking core planning skills via automatically synthesized Boolean and MCQ questions across 13 domains \\
\hline
NATURALPLAN~\citep{zheng2024natural}              
  & Natural‐Language Planning            & Real‐world NL planning tasks: Trip Planning, Meeting Planning, Calendar Scheduling         & Static dataset with in‐context tool outputs    & Evaluating LLMs on realistic NL planning: complexity variation; self‐correction; few‐shot generalization; long‐context plannin \\
\hline
FlowBench~\citep{xiao2024flowbench}               
  & Workflow Planning            & Workflow-guided planning tasks across 51 scenarios in 6 domains              & Static prompts with workflow knowledge in text, code, and flowchart formats          & Formalizing diverse workflow knowledge formats; multi-tiered evaluation (static turn-level and dynamic session-level); comparative analysis of LLM performance across formats \\
\hline
WorFBench~\citep{qiao2024benchmarking}            
  & Workflow Planning         & Graph-structured workflow generation across four scenarios: problem-solving, function calling, embodied planning, open-grounded planning  & Static dataset (task descriptions + action lists)          & Multi-scenario DAG workflow generation; evaluation via WORFEVAL’s subsequence and subgraph matching; analyzing sequence vs graph planning gap; assessing generalization and downstream task enhancement \\
\hline
MINT~\citep{wang2024mintevaluatingllmsmultiturn}  
  & Interactive Planning         &  Multi-turn interactive tool-augmented task solving with natural language feedback       & Simulated online environment (tool execution + LLM-based user feedback)           & Quantifying per-turn gains from tool use and language feedback; analyzing effects of instruction fine-tuning on multi-turn reasoning \\
\hline
REALM-Bench~\citep{geng2025realm}                
  & Interactive Planning         & Single- and multi-agent real-world scenario planning with dynamic environment feedback          & Simulator-based interactive environment                & Evaluating adaptive planning under parallelism, complex dependencies, and unexpected disruptions\\
\hline
\end{tabular}
}
\end{table*}

We categorize benchmarks of planning capabilities in Table~\ref{tab:planning benchmarks}. The columns in Table~\ref{tab:planning benchmarks} have the following meanings:
\begin{itemize}

    \item \textbf{Domain Category}: Specifies the main subject area or reasoning type for each benchmark, such as mathematical reasoning, document navigation, or scientific knowledge.
    \item \textbf{Task Type}: Describes the core task or problem-solving activity assessed by the benchmark, for example, algebraic word problem solving, multi-hop question answering, or implicit inference.
    \item \textbf{Format}: Indicates the format in which the benchmark is presented, most commonly as a static dataset containing predefined problems and questions.
    \item \textbf{Key Focus}: Summarizes the main evaluation focus of the benchmark, highlighting aspects such as rationale generation, multi-step reasoning, performance on simple or complex variations, and the incorporation of charts or images.
\end{itemize}

\subsection{\textbf{Interaction Capability}}
\subsubsection{\textbf{Static Interaction}}
\begin{table*}[htbp]
\centering
\renewcommand{\arraystretch}{1.4}
\caption{Tool Benchmark and Their Taxonomy 1}

\label{tab:tooltable1}
\resizebox{\textwidth}{!}{
\begin{tabular}{p{2cm}p{2cm}p{3.5cm}p{3.5cm}p{5cm}}
\hline
\textbf{Benchmark} & \textbf{Task Domain} & \textbf{Task Type} & \textbf{API Data Source and Authenticity} & \textbf{Instruction Data Source and Authenticity} \\ \hline
ToolAlpaca~\cite{toolalpaca} & Multiple daily life domains & Single-turn API call, Multi-turn API call & Public APIs repository on GitHub & LLMs (e.g., GPT-3.5) auto-generated tool usage dialogues \\ \hline
APIBench~\cite{apibench} & Broad machine learning tasks & Single-turn API call & PyTorch Hub, HuggingFace Official & GPT-4 self-instructed data synthesis \\ \hline
ToolBench~\cite{toolbench} & Multiple daily life domains & Single-turn API call, Multi-turn chained API calls & RapidAPI Hub & DFSDT algorithm on ChatGPT for searching and annotating \\ \hline
BFCL~\cite{BFCL} & Multiple daily life domains & Simple Function, Multiple Functions, Parallel Functions & Real usable network APIs and hand-written functions with executable feasibility & Automatically generated synthetic data \\ \hline
ToolSandbox~\cite{toolsandbox} & Mobile environment tasks & Single/Multiple Tool Call, Single/Multiple User Turn & Real usable RapidAPI interfaces and local functions & GPT-4o simulated dialogue and execution paths \\ \hline
Seal-Tools~\cite{sealtools} & Multiple daily life domains & Single-tool, multiple-tool, parallel calls, and nested calls & Real usable RapidAPI interfaces & Tool definition and example calls are synthetic \\ \hline
API-Bank~\cite{apibank} & Multiple daily life domains & Single call, Retrieval+Call, Plan+Retrieval+Call & Hand-implemented and deployed by engineering teams in the same framework & Annotated by multiple computer science students through a discussion-based method \\ \hline
NexusRaven~\cite{nexusraven} & Multiple daily life domains & Single call, parallel call, nested/composite call & Handwritten but verified for multi-turn executable feasibility & Synthetic generated, but from real open-source projects \\ \hline
API-Blend~\cite{apiblend} & Multiple daily life domains & Single API recognition, parameter extraction, multi-step API calls & LLM-assisted generated synthetic examples & 5 real or high-quality open-source benchmarks \\ \hline
RestBench~\cite{restbench} & TMDB movie database and Spotify music player & Multi-step API calls & From TMDB and Spotify official OpenAPI documentation, real and callable & 6 NLP experts designed from bottom-up using real-world usage scenarios \\ \hline
APIGen~\cite{apigen} & Multiple daily life domains & Simple Function, Multiple Functions, Parallel Functions, Parallel Multiple Functions & RapidAPI Hub real APIs & Self-guided synthesis based on models like DeepSeek-V2-Chat, Mixtral \\ \hline
StableTool Bench~\cite{stabletoolbench} & Multiple daily life domains & Single-turn API calls, Multi-turn API chain calls & From ToolBench published data, responses will consider GPT-generated API responses & Human-designed real-world instruction set \\ \hline
ComplexFunc Bench~\cite{complexfuncbench} & Booking.com-related APIs & Constraint-based multi-step calls & Synthetic generated and human-annotated through multi-turn validation & Human-designed real-world instruction set \\ \hline
NESTFUL~\cite{nestful} & Mathematical computation tools, general programming functions & Single-turn Multi-step, parameter filling & Real executable production-level functions & Manually constructed and filtered from real datasets \\ \hline
ToolEyes~\cite{tooleyes} & Seven high-demand real-world scenarios & Multi-tool calls and multi-step processes & Follows OpenAPI/JSON specification HTTP API & LangChain official repository (GitHub), SerpAPI platform, OpenAI documentation \\ \hline
PEToolBench~\cite{petoolbench} & Personalized tool calling tasks & Instruction calls with user historical data & Real RESTful APIs from RapidAPI & LLM self-synthesis and multi-turn filtering, non-real user data \\ \hline
ACEBench~\cite{chen2025acebenchwinsmatchpoint} & Multiple daily life domains & multi-step call, in-complete instruction following,  multi-agent collaboration & Synthetic APIs referenced by Real APIs & Synthetic instruction data \\ \hline
\end{tabular}}
\end{table*}

\begin{table*}[htbp]
\centering
\renewcommand{\arraystretch}{1.4}
\caption{Tool Benchmark and Their Taxonomy 2}
\label{tab:tooltable2}
\resizebox{\textwidth}{!}{
\begin{tabular}{p{2cm}p{3cm}p{3cm}p{1.3cm}p{6cm}}
\hline
\textbf{Benchmark} & \textbf{Metric} & \textbf{Evaluation Label and Method} & \textbf{Input Modality} & \textbf{Main Challenges} \\ \hline
ToolAlpaca~\cite{toolalpaca} & Procedure accuracy, Response accuracy & Label by API name + automated evaluation & Text & Generalizing to unseen real APIs using a compact language model trained with only 4K diverse simulated examples. \\ \hline
APIBench~\cite{apibench} & Accuracy, Hallucination Rate & Label by API name + automated evaluation & Text & Generalizing a compact LLM to accurately call unseen APIs in a large, overlapping, and ever-changing API set. \\ \hline
ToolBench~\cite{toolbench} & Pass Rate, Win Rate & Label by ChatGPT-generated + LLM evaluation similarity & Text & Accurate retrieval of relevant APIs and generalization to unseen APIs and complex instructions through multi-step, depth-first decision tree strategy. \\ \hline
BFCL`\cite{BFCL} & AST Evaluation, Executable Function Evaluation & Label by experts + automated evaluation similarity & Text & Generalizing to unseen functions in multi-domain, multi-language, and complex multi-step and parallel call scenarios. \\ \hline
ToolSandbox~\cite{toolsandbox} & Milestone Similarity, Turn Count & Milestones and Minefields provided by humans, automated evaluation based on DAG similarity & Text & Generalizing tool call and interaction strategies in complex dialogue scenarios with state dependency and incomplete information. \\ \hline
Seal-Tools~\cite{sealtools} & Format Accuracy, Tool Precision/Recall/F1, Parameter Precision/Recall/F1 & Label generated through LLM self-instructed multi-turn validation, automated evaluation & Text & Evaluating multi-layer nested calls and implicit dependencies, with evaluation criteria including correct parameter passing. \\ \hline
API-Bank~\cite{apibank} & ROUGE-L, Accuracy & Label by manual annotators + fully automated evaluation & Text & Generalizing LLMs to retrieve and call unknown tools accurately in multi-domain, multi-API pool, and hybrid information retrieval and planning scenarios, with real execution environment verification. \\ \hline
NexusRaven~\cite{nexusraven} & Nested/Composite Success Rate & Label manually written + automated evaluation script & Text & Handling high complexity in unseen function signatures, multi-level nesting, and parallel calls under zero-shot conditions. \\ \hline
API-Blend~\cite{apiblend} & API-F1, Parameter-F1, LCS-F1 & Label: Dataset original annotators or paper authors provide + automated evaluation scripts & Text & Cross-domain generalization in heterogeneous data and sparse sequencing requirements. \\ \hline
RestBench~\cite{restbench} & Correct Path Rate, $\Delta$ Solution Length & Label manually written + automated evaluation script & Text & Breaking down complex tasks into coarse-to-fine online planning while maintaining high robustness and low redundant calls during execution. \\ \hline
APIGen~\cite{apigen} & AST Evaluation, Executable Function Evaluation & Label by experts + automated evaluation similarity & Text & Building a pipeline for automatic data generation: “Format verification → Execution verification → Semantic verification.” \\ \hline
StableTool Bench~\cite{stabletoolbench} & Solvable Pass Rate, Solvable Win Rate & Label by three LLM votes + GPT-4 evaluation & Text & Reducing fluctuations caused by real API downtime or changes through caching and LLM simulation, and evaluating solvable tasks with GPT-4 assessment. \\ \hline
ComplexFunc Bench~\cite{complexfuncbench} & Rule-Based + Response-Based Precision/Recall/F1 & Label manually written + automated evaluation script & Text & Inferring optimal call sequences in multi-user constraint and implicit parameter scenarios while handling long parameters. \\ \hline
NESTFUL~\cite{nestful} & F1 Func/Param, Partial Sequence Accuracy & Label manually written + automated evaluation script & Text & Correctly generating function names, parameters, and managing output variables in multi-layer data dependency and variable reference nested calls.\\ 
\hline
ToolEyes~\cite{tooleyes} & Format alignment, Intent understanding, Behavior planning, Tool selection, Answer organization with average score across five dimensions & Label manually written + GPT-4 evaluation score & Text & Capturing LLM capabilities in format compliance, intent reasoning, behavior planning, tool matching, and answer integration at a fine-grained level. \\ \hline
PEToolBench~\cite{petoolbench} & Preference Alignment Improvement, Tool Call Accuracy & Label: LLM synthesized + manual writing, automated evaluation matching & Text & Identifying and selecting the best tool with user implicit preferences from tools with similar functionalities and generating precise parameters. \\ \hline
ACEBench~\cite{chen2025acebenchwinsmatchpoint} & End-to-End Accuracy, Process Accuracy & Label: Automated synthesized + expert checking, Eval: automated & Text & Achieving efficient, low-cost, and fine-grained automated evaluation of tool invocation in real-world scenarios involving multi-turn, multi-step interactions with incomplete instructions and multiple agents. \\ 
\hline
\end{tabular}
}
\end{table*}

We categorize benchmarks of static interaction capabilities in Table~\ref{tab:tooltable1} and Table~\ref{tab:tooltable2}. The columns in these tables have the following meanings:
\begin{itemize}
    \item \textbf{Task Domain}: Specifies the application areas or domains each benchmark covers, such as daily life tasks, machine learning tasks, or mobile environment tasks.
    \item \textbf{Task Type}: Describes the structure and complexity of the evaluated tasks, including single-turn or multi-turn API calls, chained or parallel functions, and multi-step tool use.
    \item \textbf{API Data Source and Authenticity}: Details the origins and authenticity of the APIs or tools involved, indicating whether they are real, public, synthetic, handwritten, or from official repositories like RapidAPI or PyTorch Hub.
    \item \textbf{Instruction Data Source and Authenticity}: Explains how the instructional or annotation data is generated for the benchmark, including whether it is LLM-generated, self-instructed, human-annotated, or derived from synthetic or real-world projects.
     \item \textbf{Metric}: Specifies the quantitative criteria or metrics used to evaluate agent performance, such as procedure accuracy, response accuracy, or hallucination rate.
    \item \textbf{Evaluation Label and Method}: Describes the way performance is measured and labeled, for example, by API name matching combined with automated evaluation procedures.
    \item \textbf{Input Modality}: Indicates the type of input data provided to the agent, such as text.
    \item \textbf{Main Challenges}: Summarizes the key difficulties or unique features faced in each benchmark, such as generalizing to unseen APIs or handling large, dynamic, and overlapping API sets.
    
\end{itemize}

\subsubsection{\textbf{Interaction with Humans}}
\begin{table*}[htbp]
\centering
\caption{Human-Agent Interaction Benchmarks and Their Taxonomy}
\label{tab:human_agent_interaction}
\resizebox{1.0\textwidth}{!}{
\begin{tabular}{p{1.5cm}p{2cm}p{2.5cm}p{2cm}p{4.3cm}p{1cm}p{2.7cm}}
\hline
\textbf{Benchmark} & \textbf{Dialogue Task Type} & \textbf{Dialogue Data Source} & \textbf{Metrics} & \textbf{Labeling and Similarity Evaluation} & \textbf{Modality} & \textbf{Explicit Interaction Evaluation Beyond Task Accuracy } \\ \hline
$\tau$-bench~\cite{yao2024tau} & Airline and retail & Simulated users via agent LLM & pass\(^k\) & Labels annotated by humans +manual similarity assessment & Text & No \\ \hline
ABCD~\cite{abcd} & 55 distinct user intents & Agent interacting with simulated users & Action State Tracking, Cascading Dialogue Success & Human annotation and manual similarity assessment & Text & No \\ \hline
MultiWOZ~\cite{budzianowski2020multiwozlargescalemultidomain} & Dialogues across multiple domains & Crowd-sourced & Belief tracking, dialogue act classification & Labels are human-provided; evaluations comparing agent responses to human benchmarks & Text & No \\ \hline
ALMITA~\cite{ALMITA} & 192 conversations for 14 intents & Combination of automated generation and manual curation & API correctness & Human annotation and automated similarity assessment & Text & No \\ \hline
IntellAgent~\cite{yao2024intellagent} & Dialogues across multiple domains & Automated process generation of diverse synthetic test cases & Task accuracy & Human annotation and automated similarity assessment & Text & No \\ \hline
\end{tabular}
}
\end{table*}

We categorize benchmarks of interaction with human capabilities in Table~\ref{tab:human_agent_interaction}. The columns in the table have the following meanings:
\begin{itemize}
    \item \textbf{Dialogue Task Type}: Describes the type or domain of dialogue tasks evaluated, such as specific user intents or multi-domain dialogues.
    \item \textbf{Dialogue Data Source}: Indicates how the dialogue data is collected or generated, e.g., via simulated users, crowd-sourcing, or automated processes.
    \item \textbf{Metrics}: Specifies the criteria or metrics used for evaluation, including task accuracy, action state tracking, belief tracking, API correctness, or dialogue success.
    \item \textbf{Labeling and Similarity Evaluation}: Explains the method for labeling responses and assessing similarity, which may involve human annotation, automated methods, or manual assessment.
    \item \textbf{Modality}: Indicates the type of input or interaction modality used, such as text.
    \item \textbf{Explicit Interaction Evaluation Beyond Task Accuracy}: States whether the benchmark includes explicit evaluation of interaction quality in addition to basic task accuracy (typically yes/no).
\end{itemize}

\subsubsection{\textbf{Interaction with Other Agents}}
\begin{table*}[ht]
\caption{Multi-Agent Benchmarks and Their Taxonomy}
\label{tab:agent-agent-interactions}
\centering
\resizebox{1.0\textwidth}{!}{
\begin{tabular}{{p{1.8cm}p{2cm}p{1cm}p{4cm}p{2cm}p{2cm}p{4cm}}}
\hline
\textbf{Benchmark} & \textbf{Interaction Paradigm} & \textbf{Modality} & \textbf{Metrics} & \textbf{Application Domain} & \textbf{Agent Architecture} & \textbf{Key Contributions} \\ \hline
BattleAgent Bench~\cite{Wang2024BattleAgentBench} & Cooperation and Competition & Text & Rule Understanding, Spatial Perception, Cooperation and Competition Score & Game & Homogeneous & Multiple difficulty levels, fine-grained evaluation \\ \hline
MultiAgent Bench~\cite{Gangrade2025MultiAgentBench} & Cooperation and Competition & Text & Task Completion, Collaboration Quality, Communication Effectiveness, Cognitive Planning & Collaborative Programming, Game Simulation & Homogeneous & Diverse communication topologies, KPI-based evaluation of individual agent contributions \\ \hline
Collab-Overcooked~\cite{Wang2025CollabOvercooked} & Cooperation & Text & Progress Completeness, Initiating Capability, Responding Capability & Game & Homogeneous & Process-oriented collaboration metrics, assessing collaborative abilities \\ \hline
Sotopia~\cite{Wang2023SOTOPIA} & Cooperation and Competition & Text & Task Completion, Social Commonsense Reasoning, Strategic Communication & Social Tasks & Homogeneous & Simulates complex social interactions among AI agents, evaluating social intelligence \\ \hline
AutoArena~\cite{Zhao2024AutoArena} & Competition & Text & Elo Ranking, Correlation with Human Preferences & Logical Reasoning, Coding Ability & Homogeneous & Swiss-system tournament and Elo ranking for evaluation \\ \hline
SmartEval\cite{Alahi2023SmartEval} & Cooperation & Text & Completion Rate, Reward, Score & Game & Homogeneous & Each game challenges different abilities, including reasoning, planning, spatial perception, etc. \\ \hline
MindAgent~\cite{Yu2024MindAgent} & Cooperation & Text & Task Completion, Coordination Efficiency, Instruction Understanding & Game & Homogeneous & Requires coordination and cooperation with human players, demonstrating planning and execution abilities \\ \hline
GOVSIM~\cite{Piatti2024GOVSIM} & Cooperation & Text & Resource Duration, Total Yield, Efficiency & Fisheries, Pastures, Pollution Management & Homogeneous & Simulates shared resource environments, requiring agents to cooperate to prevent resource depletion \\ \hline
\end{tabular}
}
\label{tab:multiagent_benchmarks}
\end{table*}

We categorize benchmarks of capabilities of interaction with other agents in Table~\ref{tab:agent-agent-interactions}. The columns in the table have the following meanings:
\begin{itemize}

    \item \textbf{Interaction Paradigm}: Specifies the nature of interactions among agents, such as cooperation, competition, or both.
    \item \textbf{Modality}: Indicates the format or channel of agent interactions, typically text-based.
    \item \textbf{Metrics}: Describes the performance metrics used for evaluation, such as rule understanding, collaboration quality, task completion, score, or resource efficiency.
    \item \textbf{Application Domain}: States the primary domain or scenario for the benchmark, such as games, collaborative programming, social tasks, or resource management.
    \item \textbf{Agent Architecture}: Indicates whether the agents in the benchmark are homogeneous (identical) or heterogeneous (diverse in capability or function).
    \item \textbf{Key Contributions}: Summarizes the unique features, innovations, or evaluation focuses of each benchmark, such as multi-level difficulty, process-oriented metrics, simulation of social intelligence, or evaluation of planning and collaboration abilities.
\end{itemize}

\subsection{\textbf{Memory Capability}}
\begin{table*}[h]
\centering   
% \vspace{-10pt}
\caption{Memory Benchmarks and Their Taxonomy}
% The symbol $\ast$ indicates a statistically significant improvement of RetrievalPRM over the best baseline with $p$-value < 0.01.}
% \vspace{-8pt}
% \hspace{-30pt}
\label{tab:memory_benchmarks}
\resizebox{1\textwidth}{!}{
\renewcommand\arraystretch{1.2}
\begin{tabular}{lp{3.5cm}p{3cm}p{3.5cm}p{2.7cm}p{3.3cm}}
\hline
Benchmark &  \textbf{Data Source}  & \textbf{Task Domain} & \textbf{Task Type} & \textbf{Memory Type}&\textbf{Metrics}\\  
\hline 

NarrativeQA~\cite{kovcisky2018narrativeqa} & Books and Movie Scripts+ Human-Generated QA Pairs  &Narrative Understanding of Books and Movies &QA &Long-range, Causal, Temporal-Spatial& BLEU-1/4, METEOR, ROUGE-L, MRR\\
\hline
QMSum~\cite{zhong2021qmsum} & Real Meeting Transcripts + Human-Annotated Queries and Summaries  &Product Design, Academic, Committee discussions&Query-Based Text Summarization&Long-range& ROUGE-1/2/L\\
\hline
QuALITY~\cite{pang2021quality}& English source articles+Human-Generated Question&Fiction, Nonfiction, Magazine&QA&Long-range, Causal& Accuracy\\
\hline
DialSim~\cite{kim2024dialsim}& TV Show Scripts + Fan QA + LLM-Generation QA&Everyday Interaction Scenes from TV Shows&QA&Long-range, Temporal-Spatial& Accuracy\\
\hline
LoCoMo~\cite{maharana2024evaluating}&LLM Generation Dialogue with Human Verification&Open-Domain Social Dialogue&QA, Event summarization, Multimodal dialogue generation&Long-range, Causal, Multimodal& F1 Score, Recall@K, ROGUE‑1/2/L, FactScore, MMRelevance \\
\hline
LTM-Benchmark~\cite{castillo2024beyond}&Synthetic conversation + Real Data&Everyday Conversations, Cognitive and Reasoning Scenarios&QA, Information Integration, Conflict Resolution&Long-range, Causal, Spatial, Foresight&Score\\
\hline
LongMemEval~\cite{wu2024longmemeval}&Manually created questions&Personalized Long-term Dialogue&Information Extraction, Multi-session Reasoning, Temporal Reasoning, Knowledge Updates, Abstention&Long-range, Dynamic Update, Temporal-Spatial&Accuracy, Recall@k, NDCG@k\\
\hline
Episodic Memory\cite{huet2025episodic}&Synthetic episodic memory data&Fictional Narrative Scenes&QA, Temporal Reasoning, Hallucination Detection&Temporal-Spatial, Long-term State Tracking&F1-Score, Kendall's \(\tau\) Coefficient, Hallucination Rate\\
\hline
PerLTQA~\cite{du2024perltqa}&LLM Generation with Human Verification&Personal Long-term Memory Simulation&QA, Memory Synthesis, Classification,  Retrieval&Semantic, Episodic&F1 Score, Accuracy, Recall@K, MAP\\
\hline
StreamBench~\cite{wu2024streambench}&Real, Human-Annotated Public Datasets&Text-to-SQL, Programming, Tool Use, Medical, Encyclopedic&Online Streaming Learning and Continuous Model Optimization&Long-range, Incremental Improvement&Accuracy, Pass@1, Exact Match\\
\hline
\end{tabular}
% \vspace{-5pt}
}
\end{table*}
We categorize benchmarks of the memory capability in Table~\ref{tab:memory_benchmarks}. The columns in the table have the following meanings:
\begin{itemize}

    \item \textbf{Data Source}: Specifies the origins of the data used in each benchmark, such as books, scripts, real meeting transcripts, synthetic conversations, or manually created questions.
    \item \textbf{Task Domain}: Describes the primary subject area or scenario being evaluated, including narrative understanding, product design, academic discussions, social dialogue, or programming.
    \item \textbf{Task Type}: Outlines the main task structure, such as question answering (QA), query-based summarization, event summarization, memory synthesis, classification, or online streaming learning.
    \item \textbf{Memory Type}: Indicates the type of memory evaluated, such as long-range, causal, temporal-spatial, semantic, episodic, foresight, multimodal, dynamic update, or incremental improvement.
    \item \textbf{Metrics}: Lists the evaluation metrics used to assess performance, for example, BLEU, ROUGE, accuracy, F1 score, NDCG@k, recall@k, MMRelevance, MAP, and hallucination rate.
\end{itemize}

% that's all folks
\end{document}